%% file: emnlp2018.tex
\documentclass[11pt,a4paper, final]{article}
\usepackage[hyperref]{emnlp2018}
\usepackage{times}
\usepackage{latexsym}
\usepackage{url}

\aclfinalcopy % Uncomment this line for the final submission

\input{imports}
\input{macros}

\newcommand{\recipeqaico}[0]{\includegraphics[width=.04\textwidth]{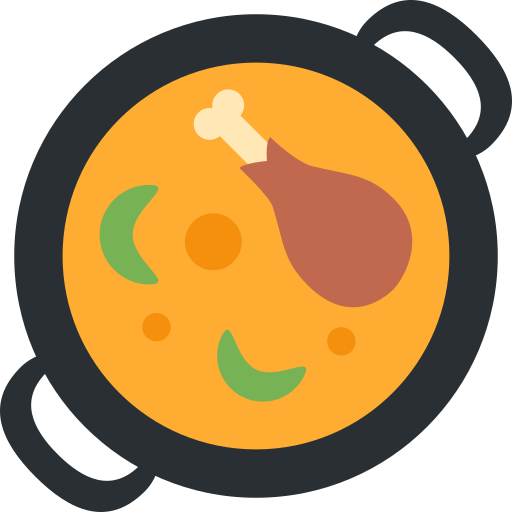}}

\input{00.title}
\date{}
\begin{document}
\maketitle

 \input{01.abstract}
 \input{01.intro}

 \input{02.dataset}
 \input{03.tasks}
 \input{04.experiments}
 \input{05.related}
 \input{06.conclusion}
 \input{07.acknowledgment}

\bibliography{emnlp2018}
\bibliographystyle{acl_natbib_nourl}

 \clearpage
 \newpage
% % \appendix
 \input{99.supp}

\end{document}

%% file: imports.tex
\usepackage{times}
\usepackage{latexsym}
\usepackage[utf8]{inputenc} % allow utf-8 input
\usepackage{url}            % simple URL typesetting
\usepackage{booktabs}       % professional-quality tables
\usepackage{amsfonts}       % blackboard math symbols
\usepackage{microtype}      % microtypography
\usepackage[centertags]{amsmath}
\usepackage{amssymb}
\usepackage{graphicx}
\usepackage{multirow}
\usepackage{xspace}
\usepackage{caption}
\usepackage[labelformat=empty]{subcaption} % or simple

\usepackage{tabularx}
\usepackage{ragged2e}
\newcolumntype{Y}{>{\RaggedRight\arraybackslash}X}
\usepackage{adjustbox}

\usepackage{xargs}  % Use more than one optional parameter in a new commands

\usepackage[most]{tcolorbox}
\usepackage{rotating}
\usepackage{enumitem}

%% file: macros.tex
\usepackage{xspace}
\makeatletter
\DeclareRobustCommand\onedot{\futurelet\@let@token\@onedot}
\def\@onedot{\ifx\@let@token.\else.\null\fi\xspace}

\def\eg{\emph{e.g}\onedot} 
\def\ie{\emph{i.e}\onedot}

\makeatother

\newcommand{\q}[1]{`#1'}

%% file: 00.title.tex
\title{\recipeqaico~RecipeQA: A Challenge Dataset for Multimodal Comprehension of Cooking Recipes}
 \author{Semih Yagcioglu, Aykut Erdem, Erkut Erdem and Nazli Ikizler-Cinbis \\
   Hacettepe University Computer Vision Lab \\
   Dept. of Computer Engineering, Hacettepe University, Ankara, TURKEY \\
   {\footnotesize{\tt semih.yagcioglu@hacettepe.edu.tr,\{aykut,erkut,nazli\}@cs.hacettepe.edu.tr}} \\
}

%% file: 01.abstract.tex
\begin{abstract}

Understanding and reasoning about cooking recipes is a fruitful research direction towards enabling machines to interpret procedural text. In this work, we introduce RecipeQA, a dataset for multimodal comprehension of cooking recipes. It comprises of approximately 20K instructional recipes with multiple modalities such as titles, descriptions and aligned set of images. With over 36K automatically generated question-answer pairs, we design a set of comprehension and reasoning tasks that require joint understanding of images and text, capturing the temporal flow of events and making sense of procedural knowledge. Our preliminary results indicate that RecipeQA will serve as a challenging test bed and an ideal benchmark for evaluating machine comprehension systems. The data and leaderboard are available at \href{http://hucvl.github.io/recipeqa}{http://hucvl.github.io/recipeqa}.

\end{abstract}

%% file: 01.intro.tex
\section{Introduction}\label{sec:intro}
There is a rich literature in natural language processing (NLP) and information retrieval on question answering (QA)~\cite{hirschman2001}, but recently deep learning has sparked interest in a special kind of QA, commonly referred to as reading comprehension (RC)~\cite{vanderwende2007answering}. The aim in RC research is to build intelligent systems with the abilities to read and understand natural language text and answer questions related to it~\cite{burges2013towards}. Such tests are appealing as they require joint understanding of the question and the related passage (\ie context), and moreover, they can analyze many different types of skills in a rather objective way~\cite{sugawara2017prerequisite}. 

Despite the progress made in recent years, there is still a significant performance gap between humans and deep neural models in RC, and researchers are pushing forward our understanding of the limitations and capabilities of these approaches by introducing new datasets. Existing tasks for RC mainly differ in two major respects: the question-answer formats, \eg cloze (fill-in-the-blank), span selection or multiple choice, and the text sources they use, such as news articles  \cite{hermann2015teaching, trischler2016newsqa}, fictional stories \cite{hill2015goldilockscbt}, Wikipedia articles \cite{narrativeqa, hewlett2016wikireading, rajpurkar2016squad} or other web sources \cite{joshi2017}. A popular topic in computer vision closely related to RC is Visual Question Answering (VQA) in which context takes the form of an image in the comprehension task, where recent datasets have also been compiled, such as \cite{antol2015vqa, yu2015visualmadlibs, johnson2016clevr, goyal2016making}, to name a few. 

\begin{figure*}[ht!]
  \footnotesize
  \begin{tabularx}{\linewidth}{@{}lY@{}l@{}Y@{}}
    \toprule

    & \textbf{Text Cloze Style Question} & \phantom{b}
                & \textbf{Context Modalities: Images and  Descriptions of Steps} \\
    \midrule
    \multicolumn{4}{@{}l}{\textbf{Recipe: Last-Minute Lasagna}} \\
    &
    \vspace{-2mm} %Ingredients: 1 24- or 26-ounce jar pasta sauce, 2 18- or 20-ounce bags refrigerated large cheese ravioli, 1 10-ounce box frozen chopped spinach, thawed and excess water squeezed out, 1 8-ounce bag shredded mozzarella, 1/2 cup (2 ounces) grated Parmesan
    \begin{enumerate}[leftmargin=*,align=left,labelwidth=\parindent,labelsep=5pt]
     \item \vspace{-2mm} Heat oven to 375 degrees F. Spoon a thin layer of sauce over the bottom of a 9-by-13-inch baking dish.
     \item \vspace{-2mm} Cover with a single layer of ravioli.
     \item \vspace{-2mm} Top with half the spinach half the mozzarella and a third of the remaining sauce. 
     \item \vspace{-2mm} Repeat with another layer of ravioli and the remaining spinach mozzarella and half the remaining sauce.
     \item \vspace{-2mm} Top with another layer of ravioli and the remaining sauce not all the ravioli may be needed. Sprinkle with the Parmesan.
     \item \vspace{-2mm} Cover with foil and bake for 30 minutes. Uncover and bake until bubbly, 5 to 10 minutes. 
    \item \vspace{-2mm} Let cool 5 minutes before spooning onto individual plates.
     \end{enumerate}
    &
    & 
    \vspace{-1mm}
     \begin{minipage}{1\linewidth}
      \includegraphics[width=0.220\linewidth]{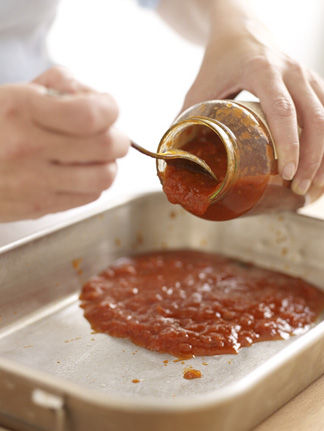}
      \includegraphics[width=0.220\linewidth]{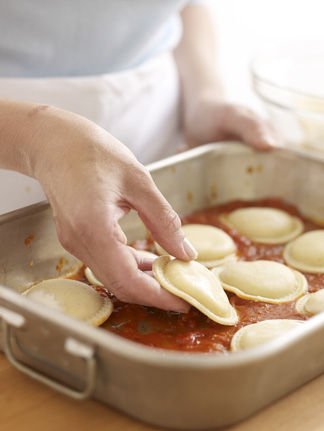}
      \includegraphics[width=0.220\linewidth]{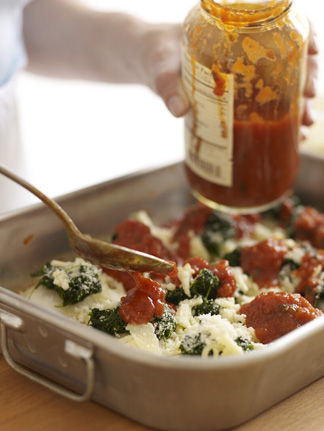} 
      \includegraphics[width=0.220\linewidth]{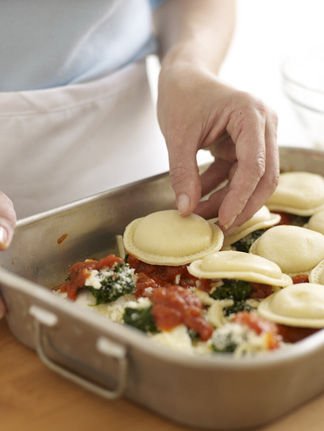}
      \begin{tabular}{@{\hspace{0.45cm}}c@{\hspace{0.975cm}}c@{\hspace{0.975cm}}c@{\hspace{0.975cm}}c}
       Step 1 & Step 2 & Step 3 & Step 4\vspace{0.375cm}
      \end{tabular}
      \includegraphics[width=0.220\linewidth]{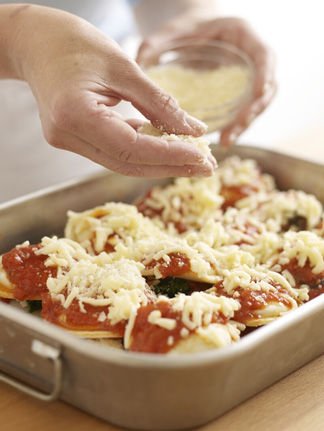}
      \includegraphics[width=0.220\linewidth]{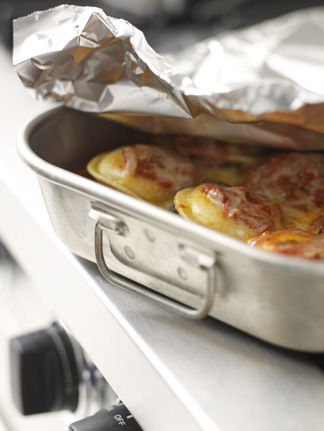}
      \includegraphics[width=0.220\linewidth]{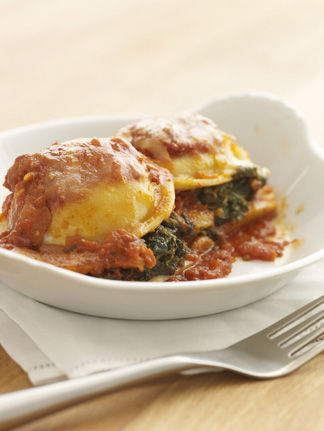}
      \begin{tabular}{@{\hspace{0.45cm}}c@{\hspace{0.975cm}}c@{\hspace{0.975cm}}c@{\hspace{0.975cm}}c}
       Step 5 & Step 6 & Step 7
      \end{tabular}
    \end{minipage}
    \\
    \midrule
    \multicolumn{4}{@{}l}{\textbf{Question} $\quad\,$ Choose the best text for the missing blank to correctly complete the recipe} \\
    \multicolumn{4}{@{}l}{ $\quad\quad\quad\qquad\;$ Cover. \hspace{0.15cm} \underline{\hspace{2cm}}. \hspace{0.15cm} Bake. \hspace{0.15cm} Cool, serve.}    \\
    \midrule
    \multicolumn{4}{@{}l}{\textbf{Answer} $\;\quad\;\,$ \textbf{A. Top, sprinkle} \hspace{0.15cm} B. Finishing touches \hspace{0.15cm} C. Layer it up \hspace{0.15cm} D. Ravioli bonus round}\\
    \bottomrule
  \end{tabularx}
  \caption{An illustrative text cloze style question (context, question and answer triplet). The context is comprised of recipe description and images where the question is generated using the question titles. Each paragraph in the context is taken from another step, as also true for the images. Bold answer is the correct one.}
  \label{fig:recipeqa}
\end{figure*}

More recently, research in QA has been extended to focus on the multimodal aspects of the problem where different modalities are being explored. \citet{tapaswi2016movieqa} introduced MovieQA where they concentrate on evaluating automatic story comprehension from both video and text. In COMICS, \citet{iyyer2016amazing} turned to comic books to test understanding of closure, transitions in the narrative from one panel to the next. In AI2D~\cite{kembhavi2016diagram} and FigureQA~\cite{kahou2017figureqa}, the authors addressed comprehension of scientific diagrams and graphical plots. Last but not least, \citet{kembhavi2017tqa} has proposed another comprehensive and challenging dataset named TQA, which comprised of middle school science lessons of diagrams and texts.

In this study, we focus on \emph{multimodal machine comprehension of cooking recipes} with images and text. To this end, we introduce a new QA dataset called \emph{RecipeQA} that consists of recipe instructions and related questions (see Fig.~\ref{fig:recipeqa} for an example text cloze style question). There are a handful of reasons why understanding and reasoning about recipes is interesting. Recipes are written with a specific goal in mind, that is to teach others how to prepare a particular food. Hence, they contain immensely rich information about the real world. Recipes consist of instructions, wherein one needs to follow each instruction to successfully complete the recipe. As a classical example in introductory programming classes, each recipe might be seen as a particular way of solving a task and in that regard can also be considered as an algorithm. We believe that recipe comprehension is an elusive challenge and might be seen as important milestone in the long-standing goal of artificial intelligence and machine reasoning~\cite{norvig1986unified,bottou14}.

Among previous efforts towards multimodal machine comprehension~\cite{tapaswi2016movieqa,kembhavi2016diagram,iyyer2016amazing,kembhavi2017tqa,kahou2017figureqa}, our study is closer to what \citet{kembhavi2017tqa} envisioned in TQA. Our task primarily differs in utilizing substantially larger number of images -- the average number of images per recipe in RecipeQA is 12 whereas TQA has only 3 images per question on average. Moreover, in our case, each image is aligned with the text of a particular step in the corresponding recipe. Another important difference is that TQA contains mostly diagrams or textbook images whereas RecipeQA consists of natural images taken by users in unconstrained environments.

Some of the important characteristics of RecipeQA are as follows:
\begin{itemize}[leftmargin=*,align=left,labelwidth=\parindent,labelsep=5pt]
\item \vspace{-2.5mm} There are arbitrary numbers of steps in recipes and images in steps, respectively.
\item \vspace{-2.5mm} There are different question styles, each requiring a specific comprehension skill.
\item \vspace{-2.5mm} There exists high lexical and syntactic divergence between contexts, questions and answers.
\item \vspace{-2.5mm} Answers require understanding procedural language, in particular keeping track of entities and/or actions and their state changes.
\item \vspace{-2.5mm} Answers may need information coming from multiple steps (\ie multiple images and multiple paragraphs).
\item \vspace{-2.5mm} Answers inherently involve multimodal understanding of image(s) and text.
\end{itemize}

To sum up, we believe RecipeQA is a challenging benchmark dataset which will serve as a test bed for evaluating multimodal comprehension systems. In this paper, we present several statistical analyses on RecipeQA and also obtain baseline performances for a number of multimodal comprehension tasks that we introduce for cooking recipes. 

%The organization of the paper is as follows. In Section \ref{sec:dataset}, we present the RecipeQA dataset. We describe a set of tasks to utilize RecipeQA in Section \ref{sec:tasks}. We describe our experimental setup in Section \ref{sec:experiments}. We discuss related work in Section \ref{sec:related}. Conclusions follow in Section \ref{sec:conclusion}.

%% file: 02.dataset.tex
\section{RecipeQA Dataset}\label{sec:dataset}

\begin{figure*}[!t]
\centering
\begin{tabular}{>{\raggedright\arraybackslash}p{0.185\linewidth}@{$\;\;$}>{\raggedright\arraybackslash}p{0.185\linewidth}@{$\;\;$}>{\raggedright\arraybackslash}p{0.185\linewidth}@{$\;\;$}>{\raggedright\arraybackslash}p{0.185\linewidth}@{$\;\;$}>{\raggedright\arraybackslash}p{0.185\linewidth}}
\scriptsize{\textbf{Intro}} & \scriptsize{\textbf{Step 1: Ingredients}} & \scriptsize{\textbf{Step 2: Prepping the Garlic, Ginger, Onion, and Tomato}} & \scriptsize{\textbf{Step 3: Drain the Chickpeas}} & \scriptsize{\textbf{Step 4: Prepping Lime and Coriander}}\\
\vspace{-1.95cm}
\includegraphics[width=0.475\linewidth]{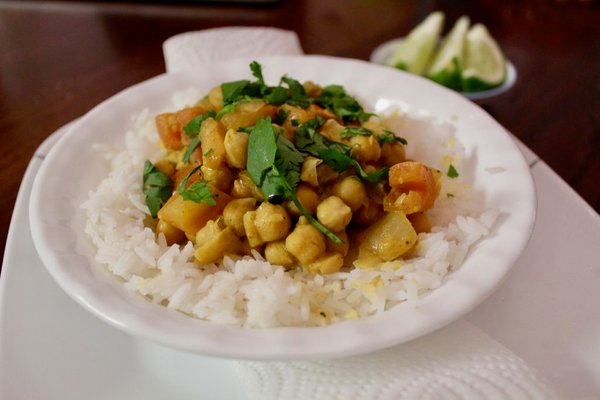} \includegraphics[width=0.475\linewidth]{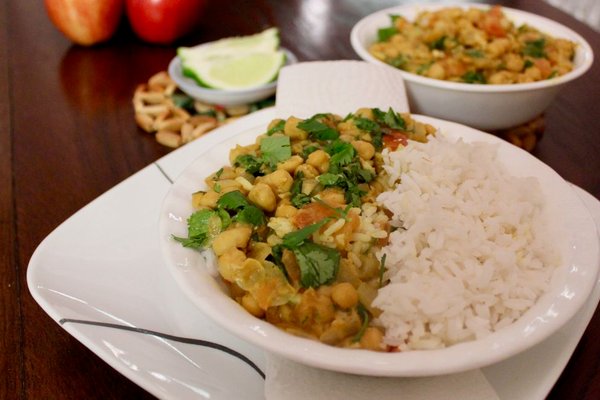} \begin{minipage}{0.185\textwidth}
\tiny{This Creamy Coconut Chickpea Curry is an quick and easy to prepare vegan and gluten free Indian-cuisine-inspired dish, made from fresh ingredients. All it takes is about 5 minutes of prep time and another 20 minutes of cooking time and you have yourself a delicious and healthy dish. Deliciously satisfying!\\}
\end{minipage} 
 &
\includegraphics[width=\linewidth]{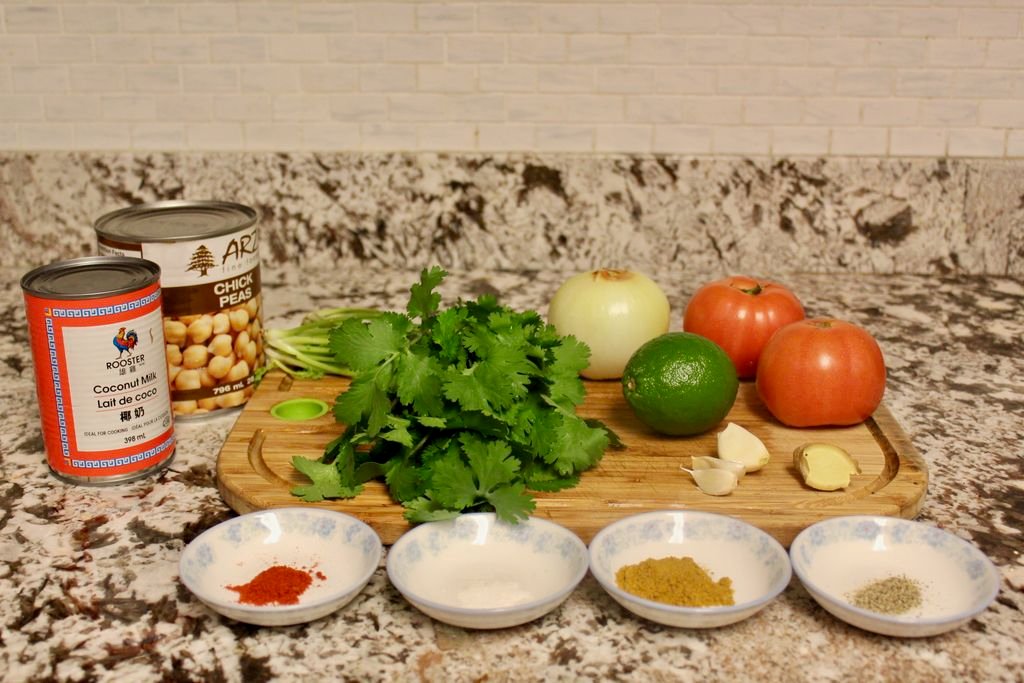}
\begin{minipage}{0.185\textwidth}
\tiny{1 can (796mL) of chickpea curry
1 can (400mL) of coconut milk
2 tomatoes
1 lime
2 stalks of coriander
3 cloves of garlic
1 inch knob of ginger
1 large yellow onion
1$/$4 teaspoon ground black pepper
1$/$2 teaspoon salt
2 teaspoon curry powder
1$/$2 teaspoon paprika
Flavourless oil like vegetable oil
Tools:
Cutting board
Knife
Skillet\\}
\end{minipage} 
 & 
\vspace{-1.95cm} \includegraphics[width=0.45\linewidth]{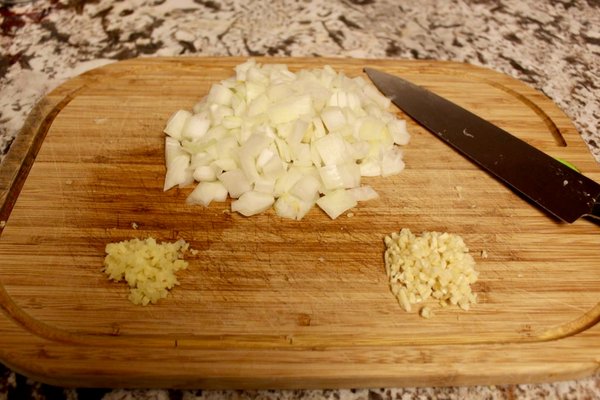} 
\includegraphics[width=0.45\linewidth]{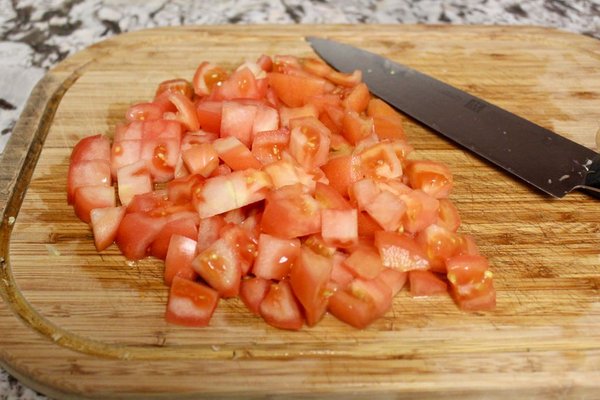} \begin{minipage}{0.185\textwidth}
\tiny{Remove the skin from the garlic, ginger, and onion. I found it easiest to use a spoon to scrape the skin from the ginger. Mince the garlic and ginger. Dice the onion
Dice the tomatoes. Once done, set aside the garlic and ginger, onions, and tomatoes on separate bowls respectively.\\}
\end{minipage} 
 & 
\includegraphics[width=\linewidth]{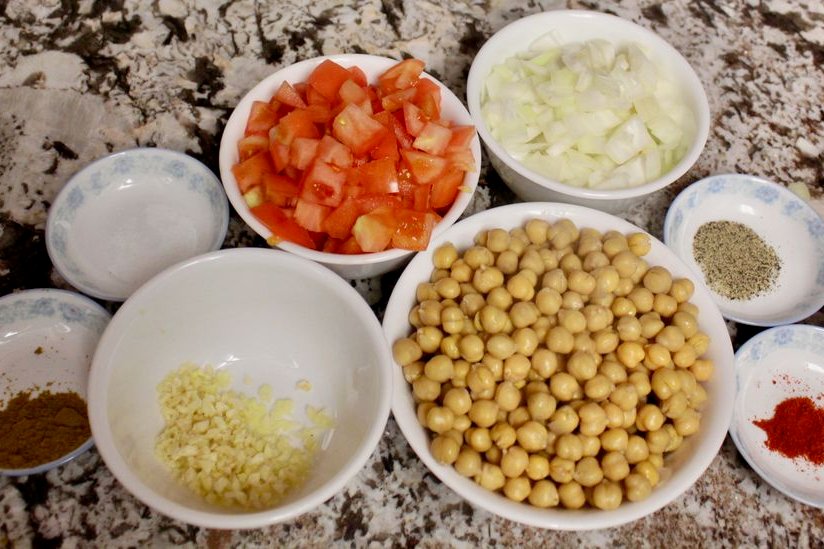}
\begin{minipage}{0.185\textwidth}
\tiny{Drain the water from the can of chickpeas. Then run rinse the chickpeas under cold water, drain very well and leave aside. My chickpeas came with the transparent outer shells of the beans, so I removed those as well, then re-rinsed it before setting it aside.\\}
\end{minipage} 
& 
\includegraphics[width=\linewidth]{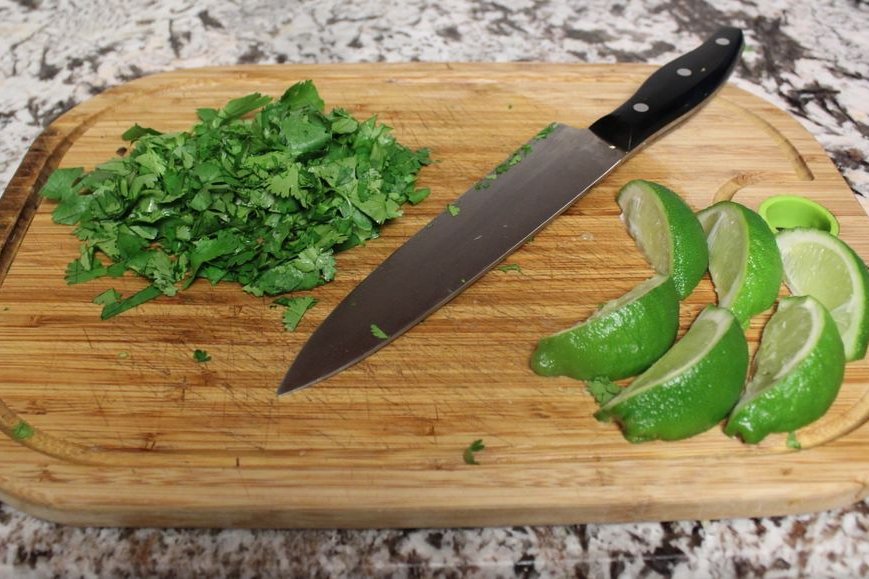} \begin{minipage}{0.185\textwidth}
\tiny{You can prep the lime and the coriander while the curry is cooking because you will have time, but I find it easier to do all the prepping at once and leave the extra time for washing the dishes. Slice the lime to 6 wedges, these will be served with the curry. Chop the leaves off from the Coriander then roughly chop it to a smaller size as it will be used for garnishing.\\}
\end{minipage}
\\
\scriptsize{\textbf{Step 5: Cook the Onion, Garlic, and Ginger}} & \scriptsize{\textbf{Step 6: Add the Spices}} & \scriptsize{\textbf{Step 7: Add the Tomatoes}} & \scriptsize{\textbf{Step 8: Add Chickpeas and Coconut Milk}} & \scriptsize{\textbf{Step 9: Garnish and Serve}}\\
\vspace{-1.95cm}
\includegraphics[width=0.633\linewidth]{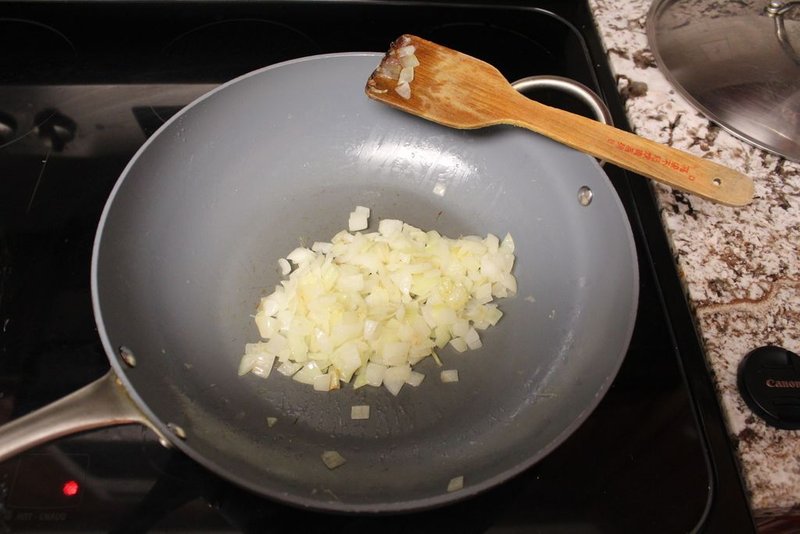} 
\includegraphics[width=0.316\linewidth]{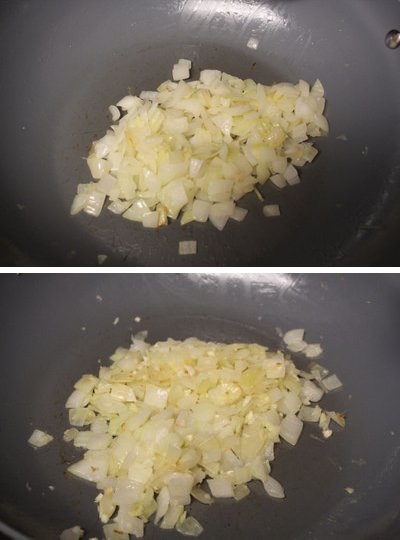}
 \begin{minipage}{0.185\textwidth}
\tiny{Heat the oil in a skillet using medium heat and add in the diced onions. Cook it until the onion softens and becomes a translucent colour. This takes around 2 to 3 minutes. Once the onions are cooked, add your garlic and ginger in and cook for another 90 seconds.\\}
\end{minipage} 
&
\vspace{-1.95cm}
\includegraphics[width=0.475\linewidth]{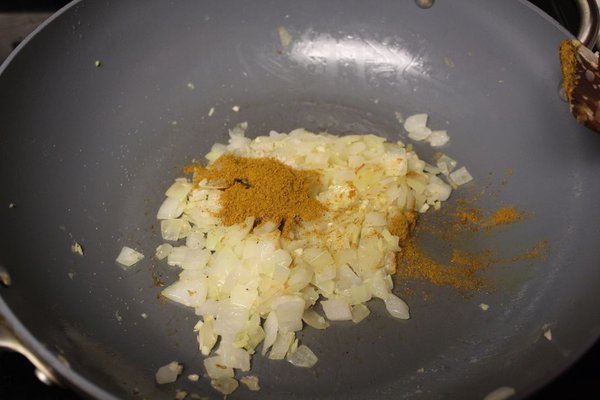} \includegraphics[width=0.475\linewidth]{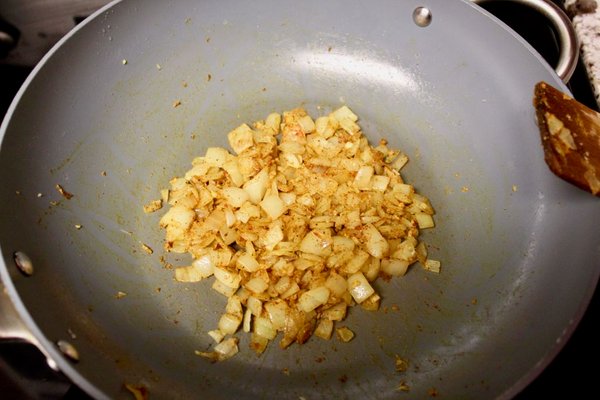} \begin{minipage}{0.185\textwidth}
\tiny{Add in all the spices (curry powder, pepper, salt, and paprika) and stir it for about 30 seconds. This will cook the spices and infuse the flavours of our spices together with the other ingredients.\\}
\end{minipage} 
&
\includegraphics[width=\linewidth]{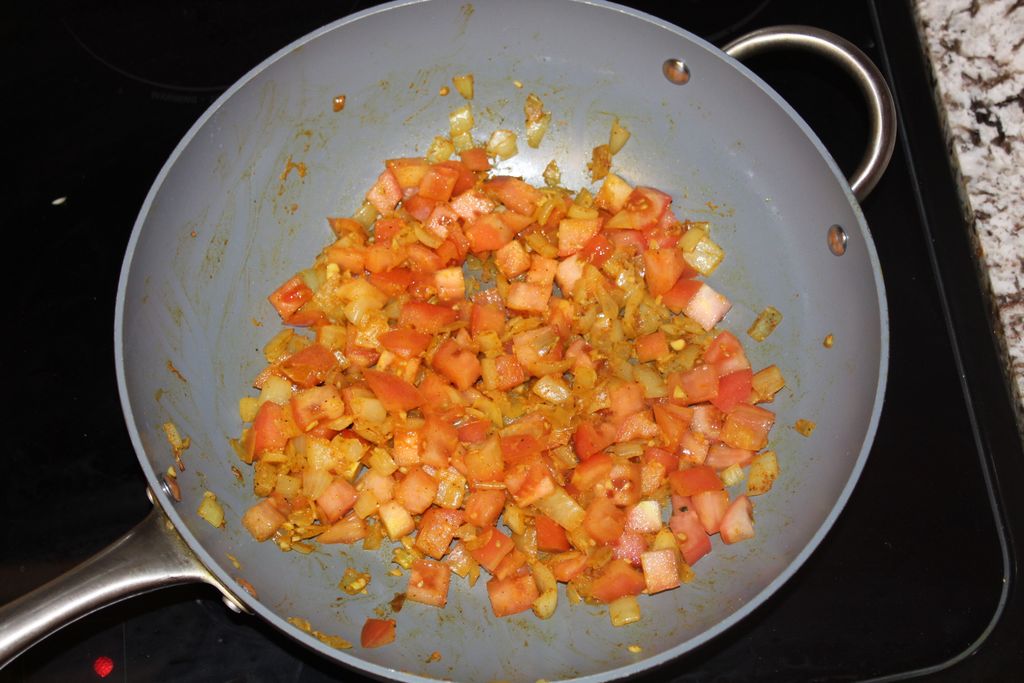}
\begin{minipage}{0.185\textwidth}
\tiny{Add the tomatoes in and stir it around until it is mixed with the spices evenly. Then leave it to cook for another 3 to 5 minutes or until the tomatoes begin break down and harden. The tomatoes add a unique texture as well as a bit of sweetness and tartness to the dish.\\}
\end{minipage}  &
\vspace{-1.95cm}
\includegraphics[width=0.475\linewidth]{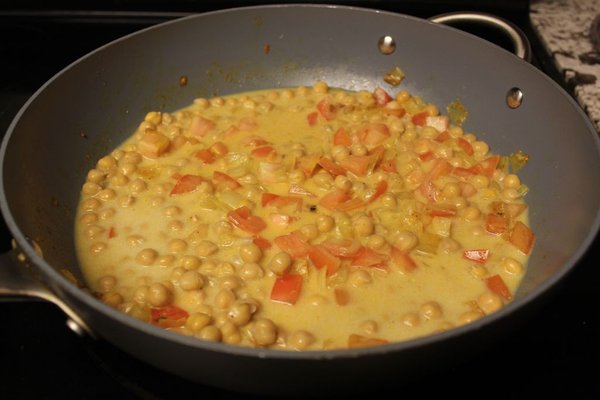} 
\includegraphics[width=0.475\linewidth]{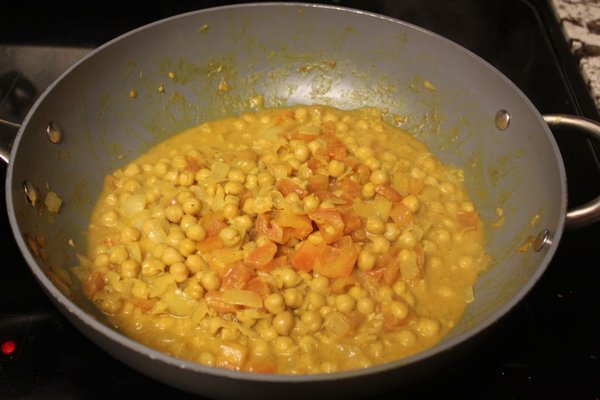} \begin{minipage}{0.185\textwidth}
\tiny{Add the drained chickpeas to the skillet with the can of coconut milk. Stir it in until the curry and the coconut milk becomes uniformly mixed. Bring the heat down to medium-low and cover it for around 15 minutes to bring it to a boil until the sauce thickens up.\\}
\end{minipage} 
 &
\vspace{-1.95cm}
\includegraphics[width=0.475\linewidth]{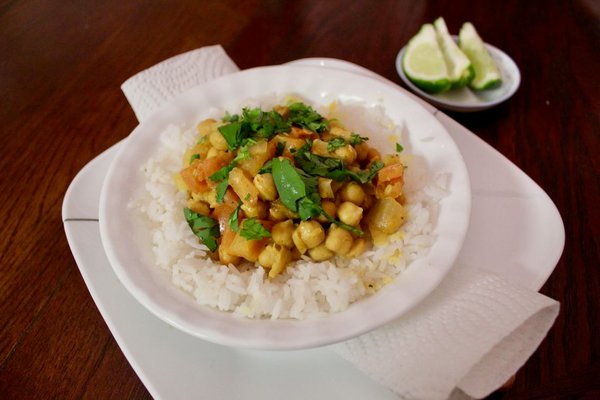} 
\includegraphics[width=0.475\linewidth]{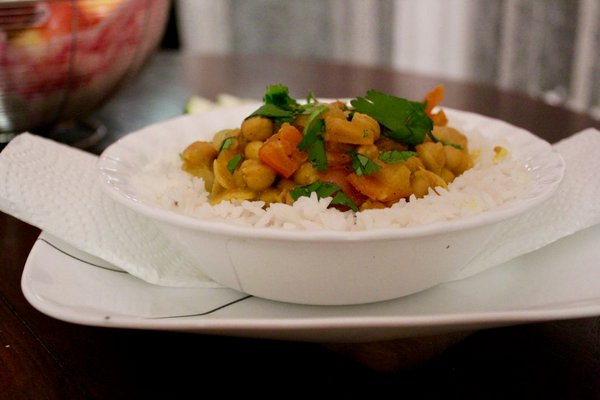} \begin{minipage}{0.185\textwidth}
\tiny{Garnish the dish with coriander and squeeze in a fresh lime on the curry to complete the dish and further elevate the flavour. Serve with with rice and enjoy!\\}
\end{minipage}
\end{tabular}
\caption[A recipe of `Creamy Coconut Chickpea Curry' recipe with 9 steps]{A recipe of  `Creamy Coconut Chickpea Curry' with 9 steps, taken from Instructables.}
\label{fig:sample-recipe}
\end{figure*}

The Recipe Question Answering (RecipeQA) dataset is a challenging multimodal dataset that evaluates reasoning over real-life cooking recipes. It consists of approximately 20K recipes from 22~food categories, and over 36K questions. Fig.~\ref{fig:sample-recipe} shows an illustrative cooking recipe from our dataset. Each recipe includes an arbitrary number of steps containing both textual and visual elements. In particular, each step of a recipe is accompanied by a \q{title}, a \q{description} and a set of illustrative \q{images} that are aligned with the title and the description. Each of these elements can be considered as a different modality of the data. The questions in RecipeQA explore the multimodal aspects of the step-by-step instructions available in the recipes through a number of specific tasks that are described in Sec.~\ref{sec:tasks}, namely \emph{textual cloze}, \emph{visual cloze}, \emph{visual coherency} and \emph{visual ordering}.

\subsection{Data Collection}
We consider cooking recipes as the main data source for our dataset. These recipes were collected from Instructables\footnote{All materials from the \href{http://www.instructables.com}{\textit{instructables.com}} were downloaded in April 2018.}, which is a how-to web site where users share all kinds of instructions including but not limited to recipes. 

We employed a set of heuristics that helped us collect high quality data in an automatic manner. For instance, while collecting the recipes, we downloaded only the most popular recipes by considering the popularity as an objective measure for assessing the quality of a recipe. Our assumption is that the mostly viewed recipes contain less noise and include easy-to-understand instructions with high-quality illustrative images. 

In total, we collected about 20K unique recipes from the food category of Instructables. We filtered out non-English recipes using a language identification~\cite{lui2012langid}, and automatically removed the ones with unreadable contents such as the ones that only contain recipe videos. Finally, as a post processing step, we normalized the description text by removing non-ASCII characters from the text.

\subsection{Questions and Answers}
\label{sec:dataset-qa}
For machine comprehension and reasoning, forming the questions and the answers is crucial for evaluating the ability of a model in understanding the content. Prior studies employed natural language questions either collected via crowdsourcing platforms such as  SQuAD~\cite{rajpurkar2016squad} or generated synthetically as in CNN/Daily Mail~\cite{hermann2015teaching}. Using natural language questions is a good approach in terms of capturing human understanding, but crowdsourcing is often too costly and does not scale well as the size of the dataset grows. Synthetic question generation is a low-cost solution, but the quality of the generated questions is subject to question. 

RecipeQA includes structured data about the cooking recipes that consists of step-by-step instructions, which helps us generate questions in a fully automatic manner without compromising the quality. Our questions test the semantics of the instructions of the recipes from different aspects through the tasks described in Sec.~\ref{sec:tasks}. In particular, we generate a set of multiple choice questions (the number of choices is fixed as four) by following a simple procedure which apply to all of our tasks with slight modifications. 

In order to generate question-answer-context triplets, we first filtered out recipes that contain less than $3$ steps or more than $25$ steps. We also ignored the initial step of the recipes as our preliminary analysis showed that the first step of the recipes almost always is used by the authors to provide a narrative, \eg why they love making that particular food, or how it makes sense to prepare a food for some occasion, and often is not relevant to the recipe instructions. In addition, we automatically removed some indicators such as step numbers that explicitly emphasize temporal order from the step titles while generating questions.

Given a task, we first randomly select a set of steps from each recipe and construct our questions and answers from these steps according to the task at hand. In particular, we employ the modality that the comprehension task is built upon to generate the candidate answers and use the remaining content as the necessary context for our questions. For instance, if the step titles are used within the candidate answers, the context becomes the descriptions and the images of the steps. As the average number of steps per recipe is larger than four, using this strategy, we can generate multiple context-question-answer triplets from a single recipe.

Candidate answers can be generated by selecting the distractors at random from the steps of other recipes. To make our dataset more challenging, we employ a different strategy and select the distractors from the relevant modalities (titles, descriptions or images), which  are not too far or too close from the correct answer. Specifically, we employ the following simple heuristic. We first find $k$ nearest neighbors ($k=100$) from other recipes. We then define an adaptive neighborhood by finding the closest distance to the query and remove the candidates that are too close. The remaining candidates are similar enough to be adversarial but not too similar to semantically substitute for the groundtruth. Finally, we randomly sample distractors from that pool. Details of the question generation procedure for each of the tasks are given in Sec.~\ref{sec:tasks}.

\subsection{Dataset Statistics}
RecipeQA dataset contains approximately 20K cooking recipes and over 36K question-answer pairs divided into four major question types reflecting each of the task at hand. The data is split into non-overlapping training, validation and test sets so that one set does not include a recipe and/or questions about that recipe which are available in other sets. There are $22$ different food categories across our dataset whose distribution is shown in Fig.~\ref{fig:recipe_categories}. While splitting the recipes into sets, we take into account these categories so that all the sets have a similar distribution of recipes across all the categories. In Table~\ref{tbl:statistics}, we show the detailed statistics about our RecipeQA dataset. Moreover, to visualize the token frequencies, we also provide the word clouds of the titles and the descriptions from the recipes in Fig.~\ref{fig:wordclouds}.

\begin{table}[!t]
\begin{center}
    {\small 
        \begin{tabular}{lrrr}
        \toprule
          & \textbf{train} & \textbf{valid} & \textbf{test}\\
          \midrule
          \vspace{0mm}
          \# of recipes                     & 15847 & 1963  & 1969  \\ % in total 19779
          \ldots avg. \# of steps           & 5.99      & 6.01 &  6.00       \\
          \ldots avg. \# of tokens (titles) & 17.79      & 17.40 &  17.67     \\
          \ldots avg. \# of tokens (descr.) & 443.01      & 440.51 &  435.33    \\
          \ldots avg. \# of images          & 12.67      & 12.74 &  12.65    \\
          \midrule
          \# of question-answers    &  29657 & 3562  & 3567     \\ % in total 36786
          \ldots textual cloze      &  7837 & 961   & 963       \\
          \ldots visual cloze       &  7144 & 842   & 848       \\
          \ldots visual coherence   &  7118 & 830   & 851       \\
          \ldots visual ordering    &  7558 & 929   & 905       \\
          \bottomrule
        \end{tabular}
    }
    \caption{RecipeQA dataset statistics.}\label{tbl:statistics}
  \end{center}
\end{table}

%% file: 03.tasks.tex
\section{Tasks}\label{sec:tasks}

RecipeQA includes four different types of tasks: (1) Textual cloze, (2) Visual cloze, (3) Visual coherence, and (4) Visual ordering. Each of these tasks requires different reasoning skills as discussed in \cite{sugawara2017prerequisite}, and considers different modalities in their contexts and candidate answer sets. By modalities, we refer to the following pieces of information: (i) titles of steps, (ii) descriptions of steps and (iii) illustrative images of steps. While generating the questions for these tasks, we rather employ fixed templates as will be discussed below, which helps us to automatically construct question-answers pairs from the recipes with no human intervention. Using these tasks, we can easily evaluate complex relationships between different steps of a recipe via their titles, their descriptions and/or their illustrative images. Hence, our question-answers pairs are multimodal in nature. In the following, we provide a detailed description of each one of these tasks and discuss our strategies while selecting candidate answers.

\subsection{Textual Cloze}
Textual cloze style questions test the ability to infer missing text either in the title or in the step description by taking into account the question's context which includes a set of illustrative images besides text. While generating the question-answer pairs for this task, we randomly select a step from the candidate steps of a given recipe, hide its title and description, and ask for identifying this text amongst the multiple choices from the remaining modalities. To construct the distractor answers, we use the strategy in Sec.~\ref{sec:dataset-qa} that depends on the WMD~\citep{kusner2015word} distance measure. In Fig.~\ref{fig:recipeqa}, we provide a sample text cloze question from RecipeQA generated automatically in this way.

\begin{figure}[!t]
\centering
  \includegraphics[width=0.95\linewidth]{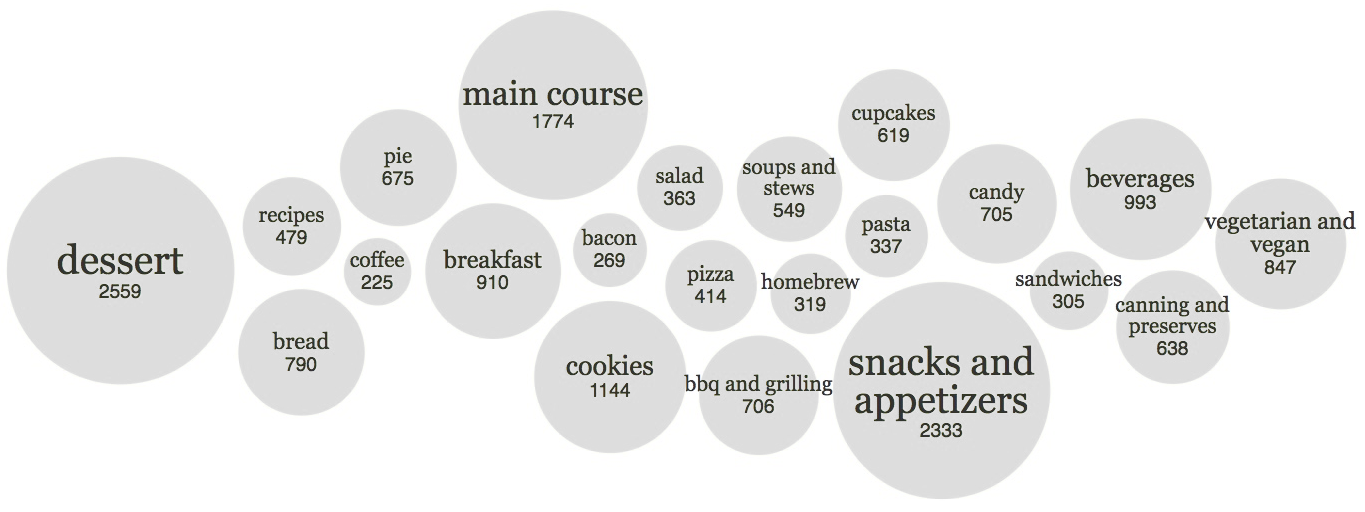} 
  \caption{Distribution of the food categories across the RecipeQA.}
  \label{fig:recipe_categories}
\end{figure}

\begin{figure}[!ht]
\centering
\begin{subfigure}[b]{.495\linewidth}
  \centering
  \includegraphics[width=1\linewidth]{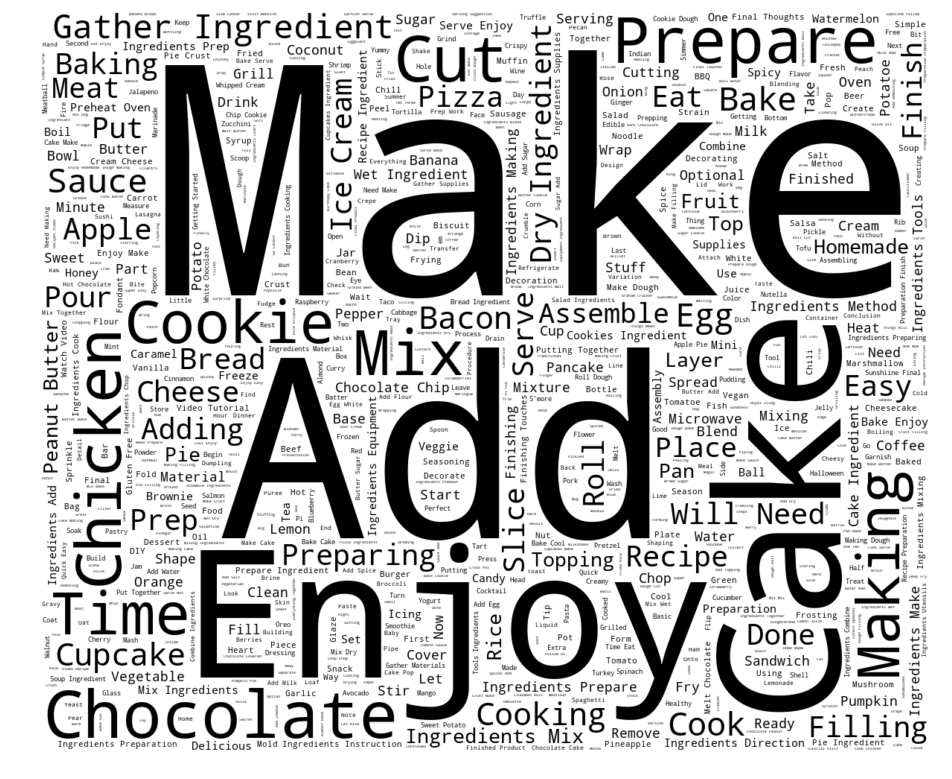}
  \caption{Step titles}
  \label{fig:wc0}
\end{subfigure}%
~
\begin{subfigure}[b]{.495\linewidth}
  \centering
  \includegraphics[width=1\linewidth]{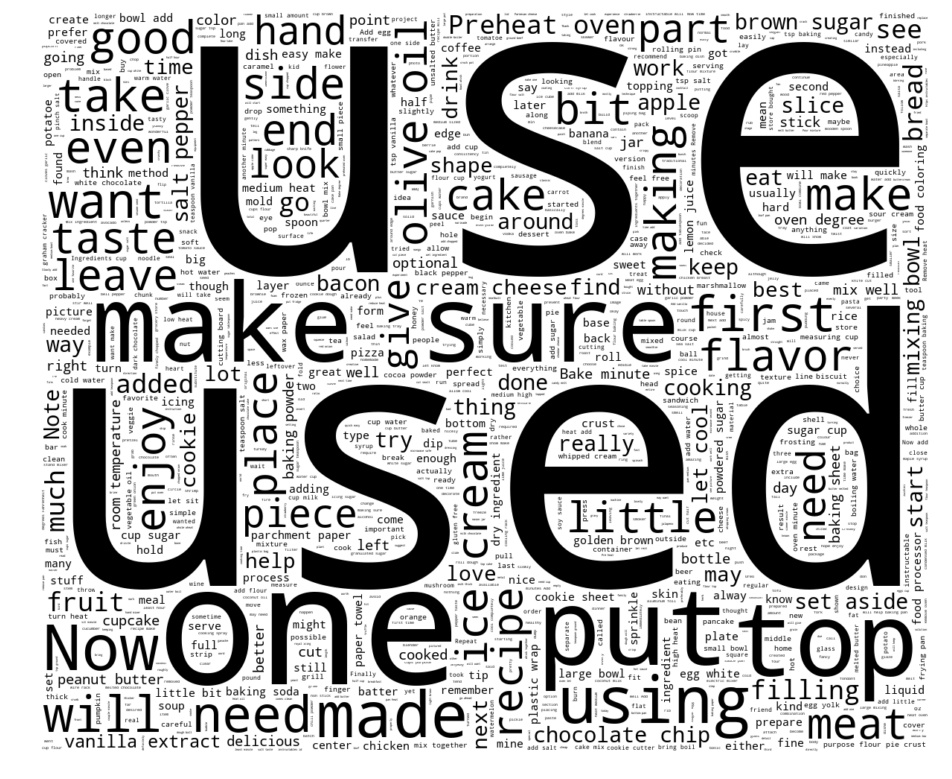}
  \caption{Step descriptions}
  \label{fig:wc1}
\end{subfigure}%
\caption[]{Word clouds of the tokens for the titles and the descriptions of the recipes from RecipeQA.}
\label{fig:wordclouds}
\end{figure}

\begin{figure*}[t!]
  \footnotesize
  \begin{tabularx}{\linewidth}{@{}l@{$\;$}Y@{}}
    \toprule
    & \hspace{7.5cm} \textbf{Context Modalities: Titles and Descriptions of Steps} \\
    \midrule
    \multicolumn{2}{@{}l}{\textbf{Recipe: Bacon Sushi}} \\
    &
    \vspace{-2mm} 
    \begin{enumerate}
     \item[\textbf{Step 1:}] \vspace{-2mm} \textbf{What You'll Need}
     This recipe makes enough bacon sushi to feed 2 - 4 people. 2 x 500g(1 lb.)  packages of bacon (I chose an applewood smoked bacon, but any type would work). 3 tbsp. oil. 1 medium onion, finely diced. 1 1\dots
     \item[\textbf{Step 2:}] \vspace{-2mm} \textbf{Cooking the Bacon} The bacon "nori" will have to be partially cooked before it can be rolled with the risotto filling. Preheat the oven to 350 degrees F. Lay half a package of bacon on the rack of the roasting pan, then bak\dots
     \item[\textbf{Step 3:}] \vspace{-2mm} \textbf{Making the Risotto Filling} I once made risotto with sushi rice, since I had no Arborio rice on hand, and I decided that the starchiness was similar in the two. My experiment was a success, and the resulting dish was just as deli\dots
     \item[\textbf{Step 4:}] \vspace{-2mm} \textbf{Jazzing Up the Risotto} Risotto is a wonderfully customizable dish, and a quick search on the internet will result in a multitude of variations. Here are two of my favorites: Asian mushroom risotto. 1 tbsp. oil. 1 package\dots
     \item[\textbf{Step 5:}] \vspace{-2mm} \textbf{Rolling the Sushi} Cover the sushi rolling mat with a large piece of aluminum foil as protection from the risotto and bacon grease. (You don't want your next sushi dinner tasting like bacon. Or maybe you do...) Lay the stri\dots
     \item[\textbf{Step 6:}] \vspace{-2mm} \textbf{Baking and Slicing} Preheat the oven to 350 degrees F. Place the aluminum foil-covered sushi rolls in the oven and bake for 20 minutes. This will warm all the ingredients and crisp the bacon a little more. It will also melt a\dots
    \item[\textbf{Step 7:}] \vspace{-2mm} \textbf{And You're Done!} Serve the sushi with a light crispy vegetable side dish, such as refreshing cucumber sticks, or a green salad. White wine makes an excellent compliment to the meal, especially if it is the same wine used in \dots
     \end{enumerate}
    \\
    \midrule
    \multirow{2}{*}{\rotatebox{90}{
        \begin{minipage}{2cm}
        \textbf{Visual Cloze\\Style Question}
        \end{minipage}}}
    & \hspace{0.65cm} \textbf{Question} $\quad$ Choose the best image for the missing blank to correctly complete the recipe \vspace{0.1cm}\\
    & 
    \hspace{2.25cm}
    \tcbset{nobeforeafter}
    \tcbox[colback=white!85!white, left=0mm,right=0mm,top=0mm,bottom=0mm,boxsep=0mm,boxrule=0.5pt]{\phantom{\includegraphics[width=0.15\linewidth]{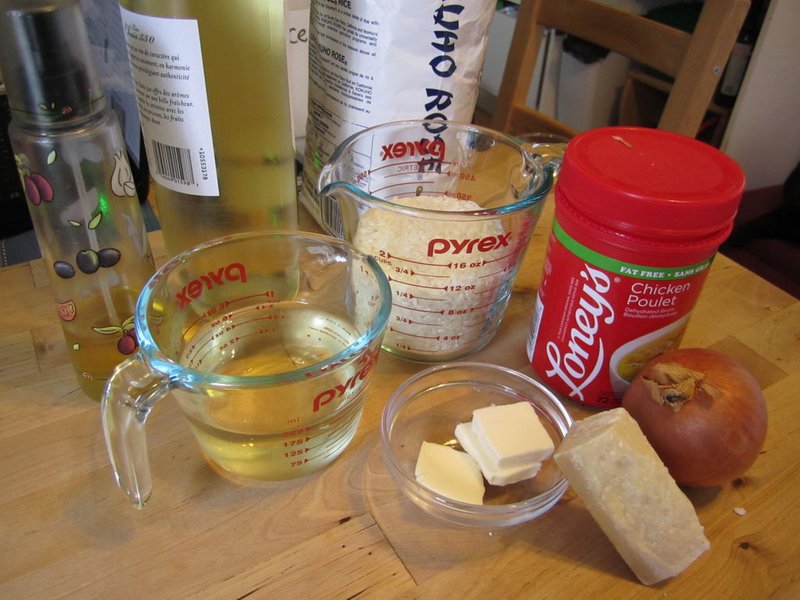}}}
    \hspace{0.25cm}
    \tcbox[colback=black!85!black, left=0mm,right=0mm,top=0mm,bottom=0mm,boxsep=0mm,boxrule=0.5pt]{\includegraphics[width=0.15\linewidth]{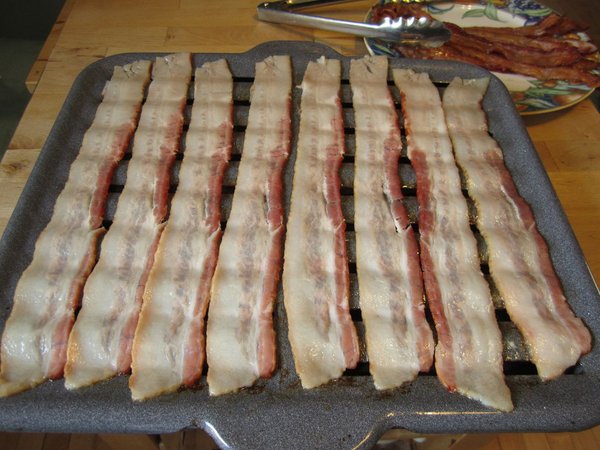}}
    \hspace{0.25cm}
    \tcbox[colback=black!85!black, left=0mm,right=0mm,top=0mm,bottom=0mm,boxsep=0mm,boxrule=0.5pt]{\includegraphics[width=0.15\linewidth]{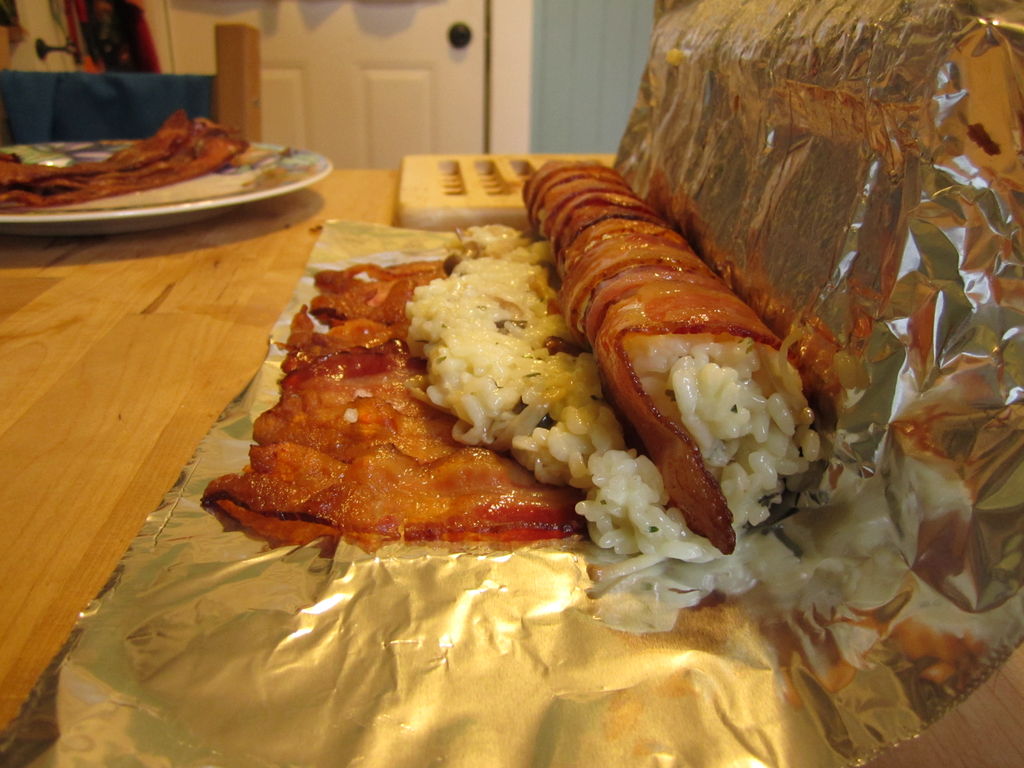}}   
    \hspace{0.25cm}
    \tcbox[colback=black!85!black, left=0mm,right=0mm,top=0mm,bottom=0mm,boxsep=0mm,boxrule=0.5pt]{\includegraphics[width=0.15\linewidth]{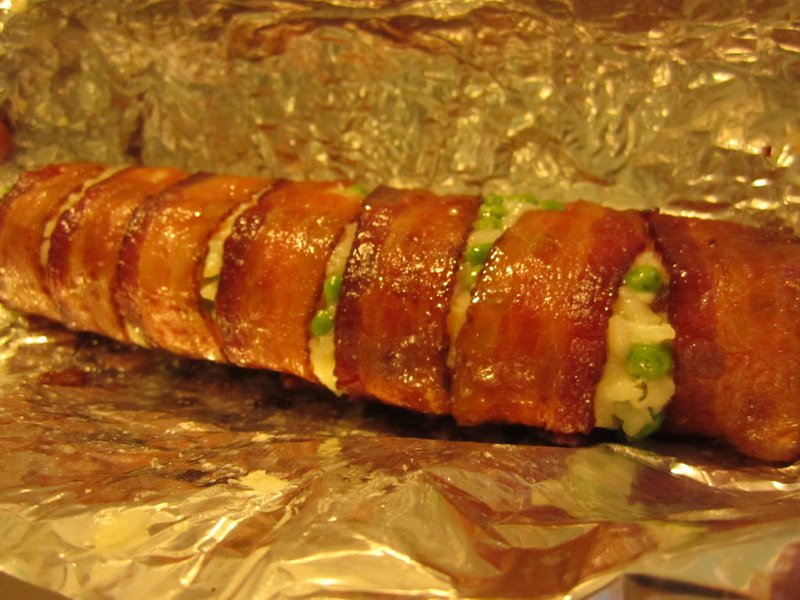}}\vspace{0.25cm}
    \\
    & \hspace{1.9cm} $\;\;\;$
    \tcbset{nobeforeafter}
    \tcbox[colback=black!85!black, left=0mm,right=0mm,top=0mm,bottom=0mm,boxsep=0mm,boxrule=0.5pt]{\includegraphics[width=0.15\linewidth]{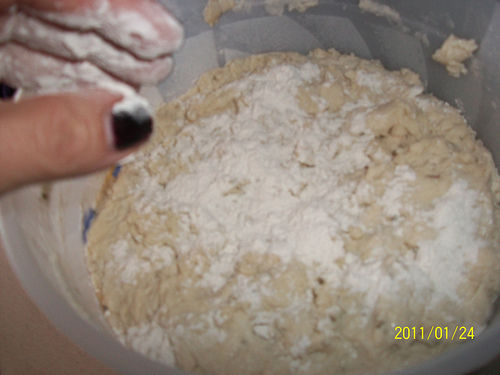}}
    \hspace{0.2cm}
    \tcbox[colframe=green!30!black,colback=green!30,left=1mm,right=1mm,top=1mm,bottom=1mm,boxsep=0mm,boxrule=0.5pt]{\includegraphics[width=0.15\linewidth]{img/bacon_sushi_1_1.jpg}}   
    \hspace{0.15cm}
    \tcbox[colback=black!85!black, left=0mm,right=0mm,top=0mm,bottom=0mm,boxsep=0mm,boxrule=0.5pt]{\includegraphics[width=0.15\linewidth]{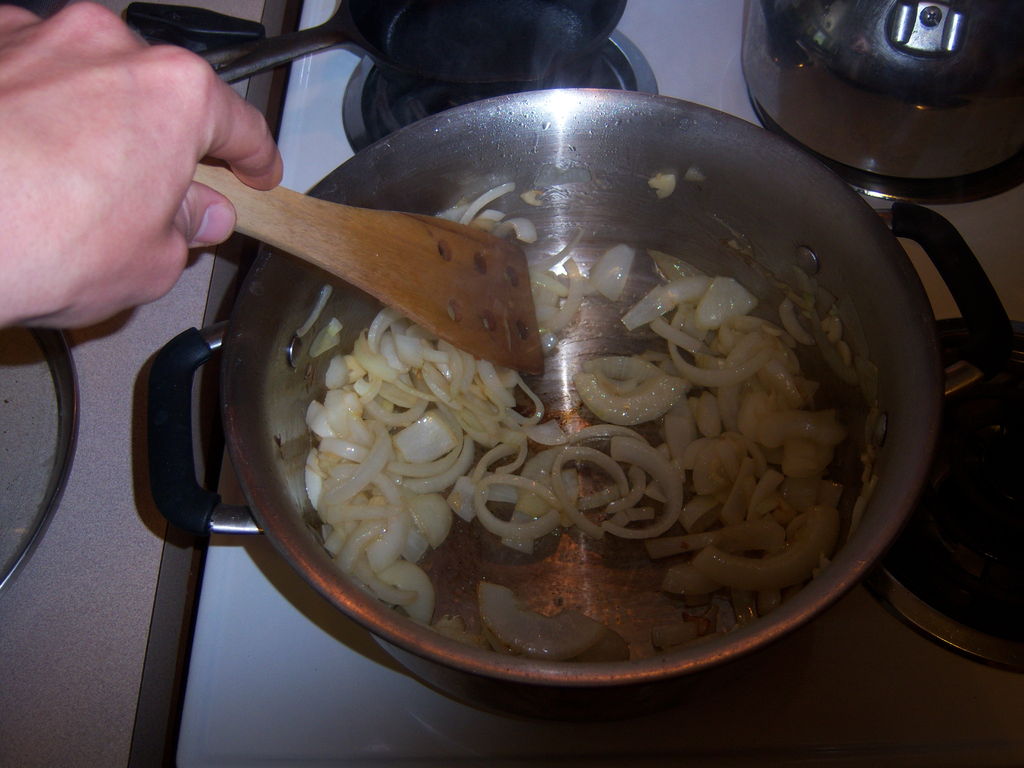}} 
     \hspace{0.2cm}
    \tcbox[colback=black!85!black, left=0mm,right=0mm,top=0mm,bottom=0mm,boxsep=0mm,boxrule=0.5pt]{\includegraphics[width=0.15\linewidth]{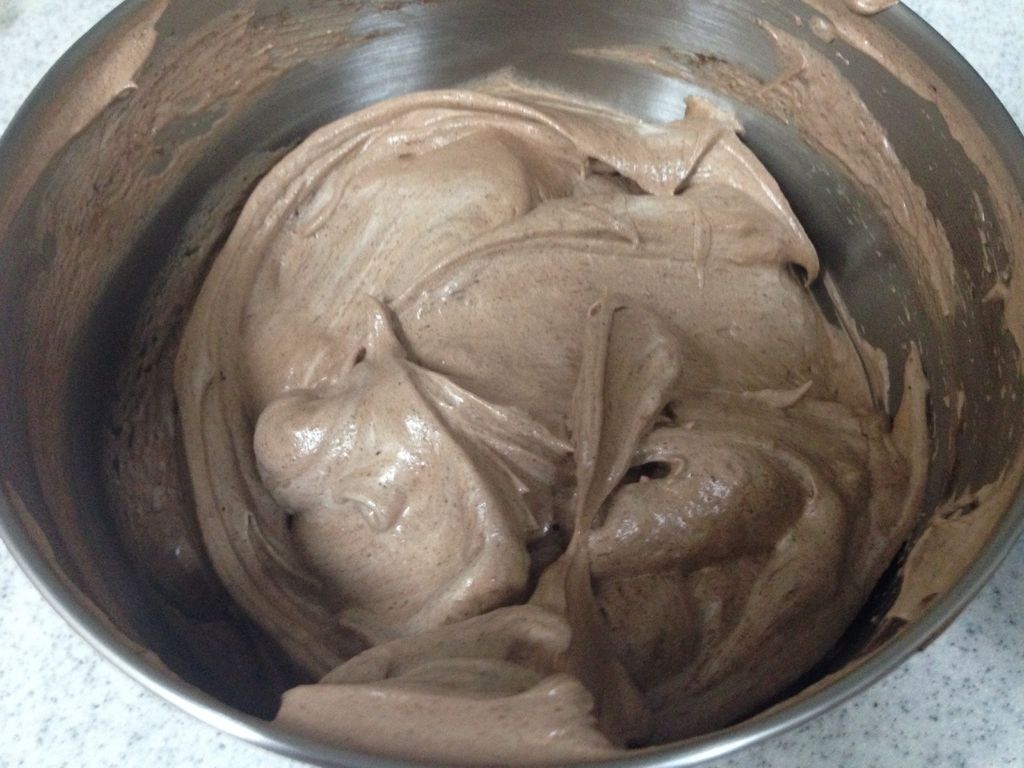}}
    \hspace{0.1cm}\\
    & \hspace{0.65cm} \textbf{Answer} \hspace{1.5cm} A. \hspace{2.3cm} \textbf{B.}
    \hspace{2.3cm} C. \hspace{2.3cm}
    D. \vspace{0.25cm} \\
    \midrule
    \multirow{2}{*}{\rotatebox{90}{
        \begin{minipage}{2.4cm}
        \textbf{Visual Coherence\\Style Question}
        \end{minipage}}}
    & \hspace{0.65cm} \textbf{Question} $\quad$ Select the incoherent image in the following sequence of images \vspace{0.1cm}\\
    & \hspace{1.9cm} $\;\;\;$
    \tcbset{nobeforeafter}
    \tcbox[colback=white!85!white, left=0mm,right=0mm,top=0mm,bottom=0mm,boxsep=0mm,boxrule=0.5pt]{\includegraphics[width=0.15\linewidth]{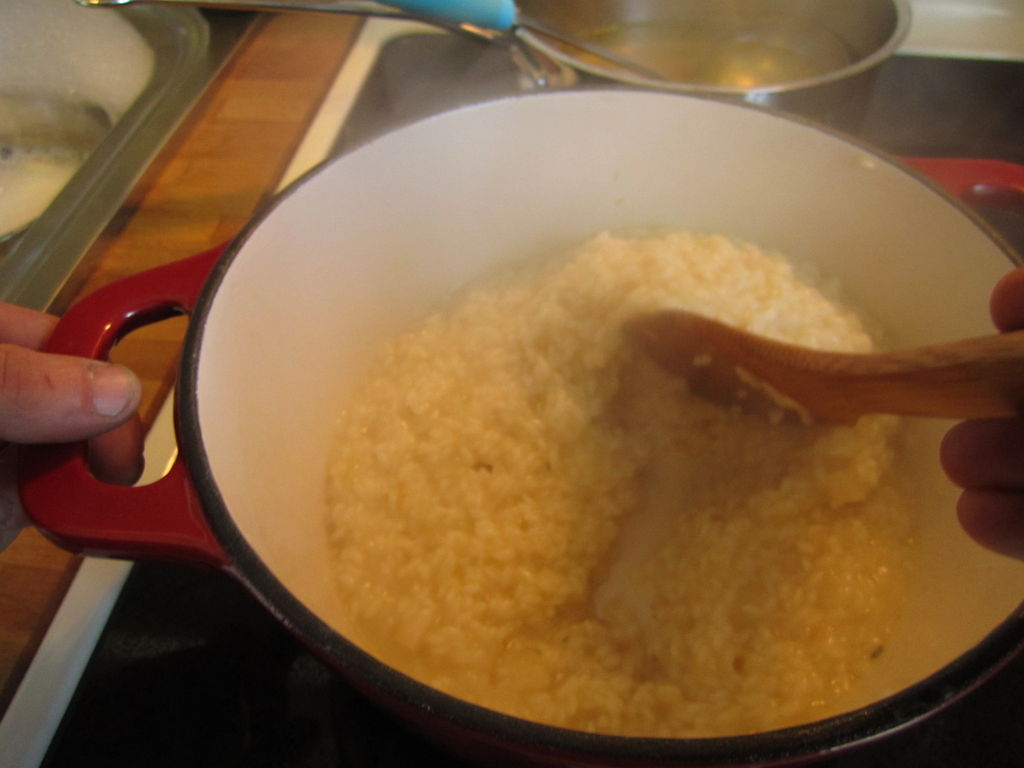}}
    \hspace{0.15cm}
    \tcbox[colframe=green!30!black,      colback=green!30,left=1mm,right=1mm,top=1mm,bottom=1mm,boxsep=0mm,boxrule=0.5pt]{\includegraphics[width=0.15\linewidth,height=0.1125\linewidth]{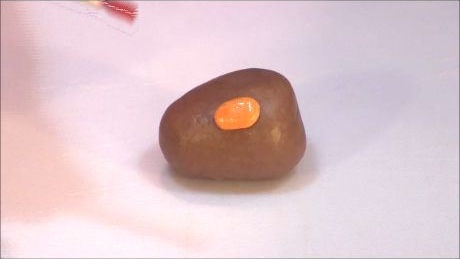}}
    \hspace{0.15cm}
    \tcbox[colback=black!85!black, left=0mm,right=0mm,top=0mm,bottom=0mm,boxsep=0mm,boxrule=0.5pt]{\includegraphics[width=0.15\linewidth]{img/bacon_sushi_5_5.jpg}}   
    \hspace{0.2cm}
    \tcbox[colback=black!85!black, left=0mm,right=0mm,top=0mm,bottom=0mm,boxsep=0mm,boxrule=0.5pt]{\includegraphics[width=0.15\linewidth]{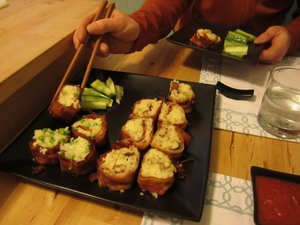}} \\
    & \hspace{0.65cm} \textbf{Answer} \hspace{1.5cm} A. \hspace{2.3cm} \textbf{B.}
    \hspace{2.3cm} C. \hspace{2.3cm}
    D. \vspace{0.25cm} \\
    \midrule
    \multirow{2}{*}{\rotatebox{90}{
        \begin{minipage}{2.225cm}
        \textbf{Visual Ordering\\Style Question}
        \end{minipage}}}
    & \hspace{0.65cm} \textbf{Question} $\quad$ Choose the correct order of the images to make a complete recipe \vspace{0.1cm}\\
    & 
    \hspace{1.9cm} $\;\;\;$
    \tcbset{nobeforeafter}
    \tcbox[colback=white!85!white, left=0mm,right=0mm,top=0mm,bottom=0mm,boxsep=0mm,boxrule=0.5pt]{\includegraphics[width=0.15\linewidth]{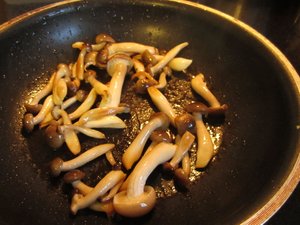}}
    \hspace{0.25cm}
    \tcbox[colback=black!85!black, left=0mm,right=0mm,top=0mm,bottom=0mm,boxsep=0mm,boxrule=0.5pt]{\includegraphics[width=0.15\linewidth]{img/bacon_sushi_7_3.jpg}}
    \hspace{0.25cm}
    \tcbox[colback=black!85!black, left=0mm,right=0mm,top=0mm,bottom=0mm,boxsep=0mm,boxrule=0.5pt]{\includegraphics[width=0.15\linewidth]{img/bacon_sushi_3_6.jpg}}   
    \hspace{0.25cm}
    \tcbox[colback=black!85!black, left=0mm,right=0mm,top=0mm,bottom=0mm,boxsep=0mm,boxrule=0.5pt]{\includegraphics[width=0.15\linewidth]{img/bacon_sushi_2_1.jpg}}
    \\
    & \hspace{3.35cm} (i) \hspace{2.3cm} (ii) \hspace{2.15cm} (iii) \hspace{2.15cm} (iv) \vspace{0.25cm} \\ 
    & \hspace{0.8cm} \textbf{Answer} $\quad$ A. (iv)-(iii)-(ii)-(i) \hspace{0.25cm} \textbf{B. (iv)-(iii)-(i)-(ii)} \hspace{0.25cm} C. (i)-(ii)-(iii)-(iv) \hspace{0.25cm} D. (ii)-(iv)-(i)-(iii) \\
    \bottomrule
  \end{tabularx}
  \caption{Sample visual cloze, visual coherence and visual ordering style questions (context, question and answer triplet) taken from the RecipeQA training set (Question Ids: 2000-3708-0-1-4-5, 3000-3708-2-3-4-6, 4000-3708-1-2-3-6). Here, the context is comprised of step titles and descriptions where the questions are generated using the images in the recipe. The correct answers are shown with green frames or in bold.}
  \label{fig:visualtasks}
\end{figure*}

\subsection{Visual Cloze}
Visual cloze style questions test a skill similar to that of textual cloze task with the difference that the missing information in this task reside in the visual domain. Here, just like the textual cloze task, for a recipe we randomly select a step, hide its representative image, and ask to infer this image amongst the multiple choices. The context for this task is all textual and is in the form of a sequence of titles and descriptions. To construct the distractor images, we use Euclidean distances of 2048-d \textit{pool5} features extracted from a ResNet-50~\cite{he2016resnet} pre-trained on ImageNet classification task. We show a sample visual cloze style question in Fig. \ref{fig:visualtasks} (second row).

\subsection{Visual Coherence}
Visual coherence style questions test the capability to identify an incoherent image in an ordered set of images given the titles and descriptions of the corresponding recipe as the context. Hence, to be successful at this task, a system needs to not only understand the relations between candidate steps, but also align and relate different modalities existing in the context and the answers. While generating the answer candidates for this task, we randomly select a single representative image from a single step and replace this image with a distractor image via employing the distractor selection strategy used for visual cloze task.  
In Fig.~\ref{fig:visualtasks} (third row), we provide a sample visual coherence style question from RecipeQA.

\subsection{Visual Ordering}
Visual ordering questions test the ability of a system in finding a correctly ordered sequence given a jumbled set of representative images of a recipe. As in the previous visual tasks, the context of this task consists of the titles and descriptions of a recipe. To successfully complete this task, the system needs to understand the temporal occurrence of a sequence of recipe steps and  infer temporal relations between candidates, \ie boiling the water first, putting the spaghetti next, so that the ordered sequence of images aligns with the given recipe. To generate answer choices, we simply use random permutations of the illustrative images in the recipe steps. In Fig.~\ref{fig:visualtasks} (last row), we illustrate this visual ordering task through an example question. Here, we should note that a similar task has been previously investigated by~\citet{agrawal2016sortstory} for visual stories where the task is to order a jumbled set of aligned image-description pairs.

%% file: 04.experiments.tex
\section{Experiments}\label{sec:experiments}

\subsection{Data Preparation}
\noindent\textbf{Ingredient Detection.} We employed the method proposed in~\cite{salvador2017learning_im2recipe} to detect recipe ingredients. To learn more effective word embeddings, we transformed the ingredients with compound words such as \textit{olive oil} into single word ingredients with a proper hyphenation as \textit{olive\_oil}.\vspace{0.1cm}

\noindent\textbf{Textual Embeddings.} We trained a distributed memory model, namely Doc2Vec~\citep{le2014distributed} and used it to learn word level and document level embeddings while encoding the semantic similarity by taking into account the word order within the provided context. In this way, we can represent each word, sentence or paragraph by a fixed sized vector. In our experiments, we employed 100-d vectors to represent all of the textual modalities (titles and descriptions). We made sure that the embeddings encode semantically useful information by exploring nearest neighbors (see Fig.~\ref{fig:doc2vec} for some examples.)

\begin{figure}[!htb]
    \centering
    \begin{tabularx}{\linewidth}{p{3.6cm}@{$\;\;\;$}p{3.6cm}}
    \hline
    \small{\textbf{Query}} & \small{\textbf{Nearest Neighbor}}\\
         \hline
         \footnotesize{Then add the green onion and garlic.} & 
         \footnotesize{Then add the white onion, red pepper and garlic.}\\
         \footnotesize{It will thicken some while it cools} &
         \footnotesize{Some cornflour to thicken.}\\
         \footnotesize{Slowly whisk in the milk, scraping the bottom and sides with a heatproof spatula to make sure all the dry ingredients are mixed in.} &
         \footnotesize{Stir the dry ingredients in, incrementally, mixing on low speed and scraping with a spatula after each addition.}\\
         \hline
    \end{tabularx}
    \caption{Sample nearest neighbors from the embeddings by the trained Doc2Vec model.}
    \label{fig:doc2vec}
\end{figure}

\noindent\textbf{Visual Features.} We used the final activation of the ResNet-50~\cite{he2016resnet} model trained on the ImageNet dataset~\cite{Russakovsky2015imagenet} to extract 2048-d dense visual representations. Then, we further utilized an autoencoder to decrease the dimension of the visual features to 100-d so that they become compatible in size with the text embeddings.

\begin{table*}[!htb]
\centering
\begin{tabular}{lcccc}
\hline
& Visual & Textual & Visual & Visual \\
& Cloze  & Cloze  & Coherence & Ordering \\ \hline
Hasty Student                  & 27.35          & 26.89          & \textbf{65.80}   & \textbf{40.88}  \vspace{0.1cm}\\
Impatient Reader (Text only)   & --              & 28.03          & --                & --               \\
Impatient Reader (Multimodal) & \textbf{27.36} & \textbf{29.07} & 28.08            & 26.74           \\
                               \hline
\end{tabular}
\caption{Results for simple and neural models on the test set of RecipeQA dataset.}
\label{tbl:results}
\end{table*}

\subsection{Baseline Models}\label{sec:baseline}
\noindent\textbf{Neural Baselines.} For our neural baselines, we adapted the Impatient Reader model in~\citep{hermann2015teaching}, which was originally developed only for the cloze style text comprehension questions in the CNN/Daily Mail dataset. In our implementation, we used a uni-directional stacked LSTM architecture with $3$ layers, in which we feed the context of the question to the network in a sequential manner. Particularly, we preserve the temporal order of the steps of the recipe while feeding it to the neural model, by mimicking the most common reading strategy -- reading from top to bottom. For the multimodal setting, since images are represented with vectors which are of the same size with the text embeddings, we also feed the images to the network in the same order they are presented in the recipe.

In order to account for different question types, we employ a modular architecture, which requires small adjustments to be made for each task. For instance, we place the candidate answers into query for the cloze style questions or remove the candidate answer from the query for the visual coherence type questions. In training our Impatient Reader baseline model, we use a cosine similarity function and employed the \emph{hinge ranking loss}~\citep{collobert2011jmlr} as follows:
\begin{equation}
L = \max \{ 0, M-cos(q, a_+)+cos(q, a_-) \}
\end{equation}
\noindent where $M$ is a scalar denoting the margin, $a_+$ represents the ground truth answer, and $a_-$ corresponds to an incorrect answer which is sampled randomly from the whole answer space. For all of our experiments, we select $M$ as $1.5$ and employ a simple heuristic to prevent overfitting by following an early stopping scheme with patience set to $10$ against the validation set accuracy after the initial epoch. For the optimizer, we use ADAM and set the learning rate to $1e-3$. The training took around $18$ to $24$ hours on GTX 1080Ti on a single GPU. We did not perform any hyperparameter tuning. \vspace{0.15cm}

\noindent\textbf{Simple Baselines.} We adapt the Hasty Student model described in \cite{tapaswi2016movieqa}, which does not consider the provided context and simply answers questions by only looking at the similarities or the dissimilarities between the elements in questions and the candidate answers. 

For the textual close task, each candidate answer is compared against the titles or descriptions of the steps by using WMD~\cite{kusner2015word} distance, where such distances are averaged. Then, the choice closest to all of the question steps is selected as the final answer. For the visual cloze task, a similar approach is carried out by considering images instead of text using deep visual features. For the visual coherence task, since the aim is to find the incoherent image among other images, the final answer is chosen as the most dissimilar one to the remaining images on average. Lastly, for the visual ordering task, first, the distances between each consecutive image pair in a candidate ordering of the jumbled image set is estimated. Then, each candidate ordering is scored based on the average of these pairwise distances and the choice with the minimum average distance is set as the final answer. In all these simple baseline models, we use the cosine distance to rank the candidates.

\subsection{Baseline Results}
We report the performance of the baseline models in Table~\ref{tbl:results} which indicates the ratio of correct answers against the total questions in the test.
For the textual cloze, the comparison between text-only and multimodal Impatient Reader models shows that the additional visual modality helps the model to understand the question better and to provide more accurate answers. While for the cloze style questions, the Impatient Reader outperforms the Hasty student, for the visual coherence and visual ordering style questions Hasty student gives way better results. This demonstrates that better neural models are needed to be able to effectively deal with this kind of questions. Some qualitative examples are provided in the supplementary material.

%% file: 05.related.tex
\section{Related Work}\label{sec:related}

Question Answering has been studied extensively in the literature. With the success of deep learning approaches in question answering, comprehension and reasoning aspects of the task has attracted  researchers to investigate QA as a medium to measure intelligence. Various datasets and methods have been proposed for measuring different aspects of the comprehension and reasoning problem. Each dataset has its own merits as well as weaknesses. Recently, a thorough analysis by \cite{chen2016thorough} revealed that the required reasoning and inference level was quite simple for CNN/Daily Mail dataset \citep{hermann2015teaching}. To make reasoning task more realistic, new datasets such as SQuAD \citep{rajpurkar2016squad}, NewsQA \citep{trischler2016newsqa}, MSMARCO \citep{nguyen2016msmarco}, CLEVR \citep{johnson2016clevr}, COMICS \citep{iyyer2016amazing} and FigureQA \citep{kahou2017figureqa} have been proposed.

In the following, we briefly discuss the publicly available datasets that are closely related to our problem and provide an overview in Table \ref{tbl:datasetcomparison}. 

The closest works to ours are \cite{iyyer2016amazing}, \cite{tapaswi2016movieqa} and \cite{kembhavi2017tqa} where data multi-modality is the key aspect. 
COMICS dataset \cite{iyyer2016amazing} focus on comic book narratives and explore visual cloze style questions, introducing a dataset consisting of drawings from comic books. The dataset is constructed from 4K Golden Age (1938-1954) comic books from the Digital Comics Museum and contains 1.2M panels with 2.5M textboxes. Three tasks are evaluated in this context, namely text cloze, visual cloze, character coherence.
\begin{table}[!t]
\centering
\resizebox{\linewidth}{!}{
\begin{tabular}{lrrl}
\hline
Dataset        & \#Images     & \#Questions & Modality         \\ \hline
COMICS         & 1.2M  & 750K       & Image/Text       \\
MovieQA        & 408  & 14,944      & Image/Video/Text \\
TQA            & 3,455        & 26,260      & Image/Text       \vspace{0.15cm}\\
RecipeQA      & 250,730      & 36,786             & Image/Text            \\
\hline
\end{tabular}
}
\caption{Comparison of the RecipeQA dataset to other multimodal machine comprehension datasets.}
\label{tbl:datasetcomparison}
\end{table}
MovieQA dataset \cite{tapaswi2016movieqa}, comprises of 15K crowdsourced questions about 408 movies. It consists of movie clips, subtitles, and snapshots, is about comprehending stories about movies.  
TQA dataset \cite{kembhavi2017tqa}, have 26K questions about 1K middle school science lessons with 3.5K images, mostly of diagrams and aims at addressing middle school knowledge acquisition using both images and text. Since the audience is middle school children, it requires limited reasoning.

RecipeQA substantially differentiates from the previous work in the following way. Our dataset consists of natural images that are taken by anonymous users in unconstrained environments, which is a major diversion from COMICS and TQA datasets. 

It should also be noted that there has been a long history of research involving cooking recipes. Recent examples include parsing of recipes ~\cite{malmaud2014,jermsurawong2015}, aligning instructional text to videos~\cite{malmaud2015s,sener2015}, recipe text generation~\cite{kiddon2016}, learning cross-modal embeddings~\cite{salvador2017learning_im2recipe}, tracking entities and action transformations in recipes~\cite{bosselut2018}.

Finally, to our best knowledge, there is no dataset focusing on ``how-to'' instructions or recipes; hence, this work will be the first to serve multimodal comprehension of recipes having an arbitrary number of steps aligned with multiple images and multiple sentences.

%% file: 06.conclusion.tex
\section{Conclusion}\label{sec:conclusion}

We present RecipeQA, a dataset for multimodal comprehension of cooking recipes, which consists of roughly 20K cooking recipes with over 36K context-question-answer triplets. To our knowledge, RecipeQA is the first machine comprehension dataset that deals with understanding procedural knowledge in a multimodal setting. Each one of the four question styles in our dataset is specifically tailored to evaluate a particular skill and requires connecting the dots between different modalities. Results of our baseline models demonstrate that RecipeQA is a challenging dataset and we make it publicly available for other researchers to promote the development of new methods for multimodal machine comprehension. In the future, we also intend to extend the dataset by collecting natural language questions-answer pairs via crowdsourcing. We also hope that RecipeQA will serve other purposes for related research problems on cooking recipes as well.

%% file: 07.acknowledgment.tex
\section*{Acknowledgments}
We would like to thank our anonymous reviewers for their insightful comments and suggestions, which helped us improve the paper, Taha Sevim and Kenan Hagverdiyev for their help in building the RecipeQA challenge website, and NVIDIA Corporation for the donation of GPUs used in this research. This work was supported in part by a Hacettepe BAP fellowship (FBB-2016-11653) awarded to Erkut Erdem. Semih Yagcioglu was partly sponsored by STM A.Ş.

%% file: 99.supp.tex
\section{Supplementary Notes}
%\vspace{-28cm}
\appendix
In the following we provide a few prediction results from the baseline models for each task.

\begin{figure*}[t!]
  \footnotesize
  \begin{tabularx}{\linewidth}{@{}l@{$\;$}Y@{}}
    \toprule
    & \hspace{7.5cm} \textbf{Context Modalities: Images and Descriptions of Steps} \\
    \midrule
    \multicolumn{2}{@{}l}{\textbf{Recipe: Grans-Green-Tomato-Chutney}} \\
    &
    \vspace{-2mm} 
    \begin{enumerate}
     \item[\textbf{Step 1:}] \vspace{-2mm} 
     
     Ingredients: 2.5kg green tomatoes, roughly chopped 0.5kg onions, finely sliced 4 tsp / 30g salt1L malt vinegar 0.5kg soft light brown sugar 250g sultanas, 1 1\dots
     
     \tcbset{nobeforeafter}
      \tcbox[colback=black!85!black, left=0mm,right=0mm,top=0mm,bottom=0mm,boxsep=0mm,boxrule=0.5pt]{\includegraphics[width=0.15\linewidth]{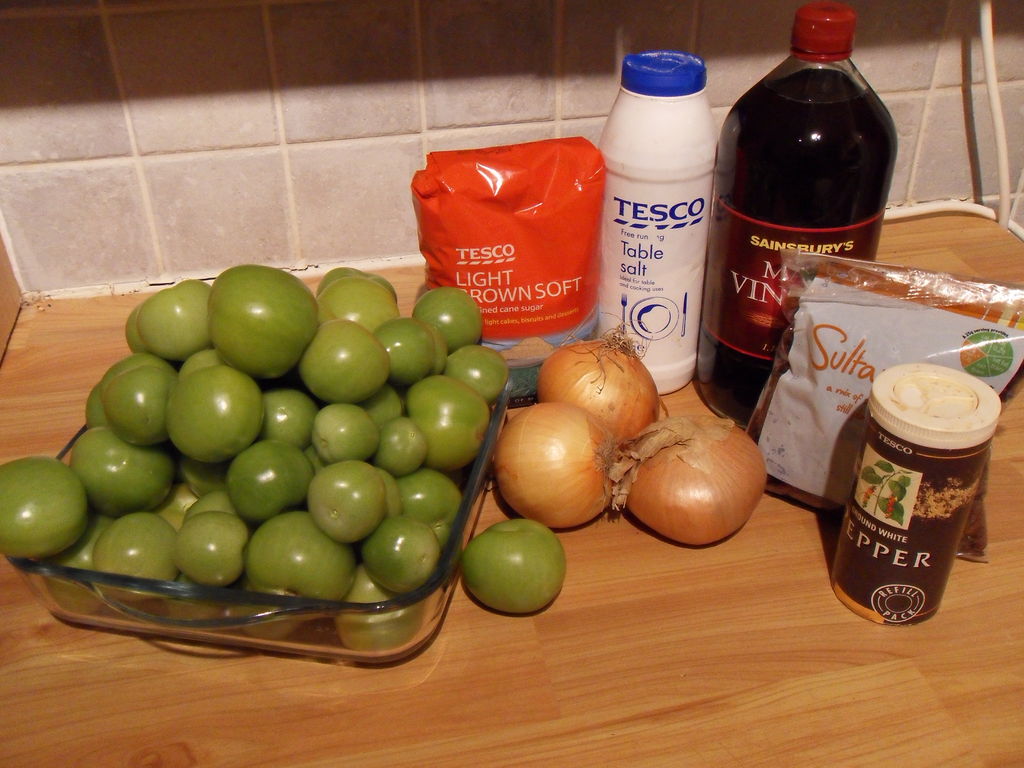}}
     
     \item[\textbf{Step 2:}]  Finely slice your onions and washed green tomatoes, cutting out any bad bits. Add to a large bowl and stir. Add the 4 teaspoons of salt, stir again and then cover with food wrap or a large plate and leave overnight.This will draw out lots of the tomato juices and help enhance the flavours.\dots
     
     \tcbset{nobeforeafter}
     \tcbox[colback=black!85!black, left=0mm,right=0mm,top=0mm,bottom=0mm,boxsep=0mm,boxrule=0.5pt]{\includegraphics[width=0.15\linewidth]{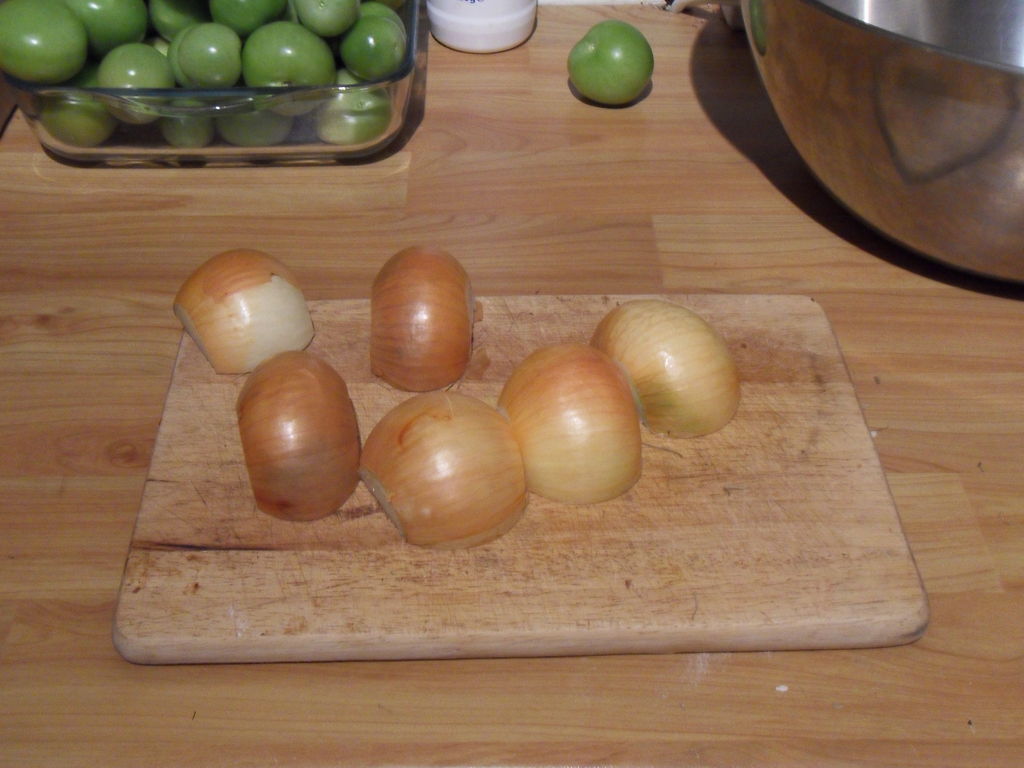}}
     \tcbox[colback=black!85!black, left=0mm,right=0mm,top=0mm,bottom=0mm,boxsep=0mm,boxrule=0.5pt]{\includegraphics[width=0.15\linewidth]{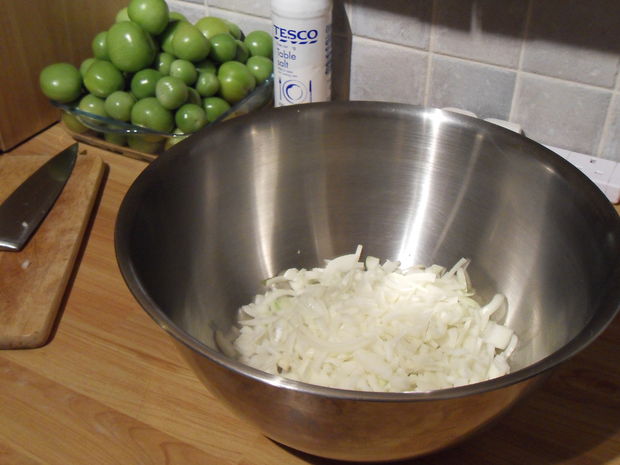}}
     \tcbox[colback=black!85!black, left=0mm,right=0mm,top=0mm,bottom=0mm,boxsep=0mm,boxrule=0.5pt]{\includegraphics[width=0.15\linewidth]{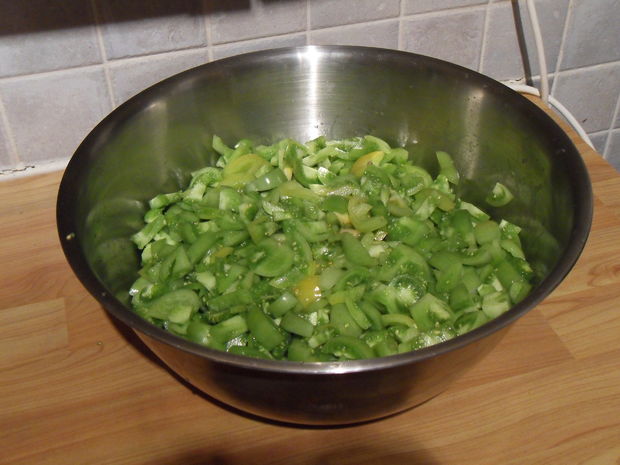}}
     \tcbox[colback=black!85!black, left=0mm,right=0mm,top=0mm,bottom=0mm,boxsep=0mm,boxrule=0.5pt]{\includegraphics[width=0.15\linewidth]{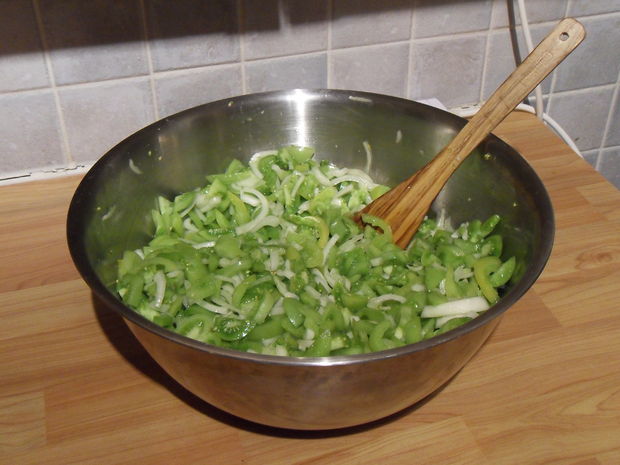}}
    
    \item[\textbf{Step 3:}] \vspace{-2mm} The next day...Place the litre of vinegar into a large pan. Add the 500g of light brown soft sugar and stir over a medium heat until all the sugar has dissolved.Bring to the boil.\dots
    
     \tcbset{nobeforeafter}
     \tcbox[colback=black!85!black, left=0mm,right=0mm,top=0mm,bottom=0mm,boxsep=0mm,boxrule=0.5pt]{\includegraphics[width=0.15\linewidth]{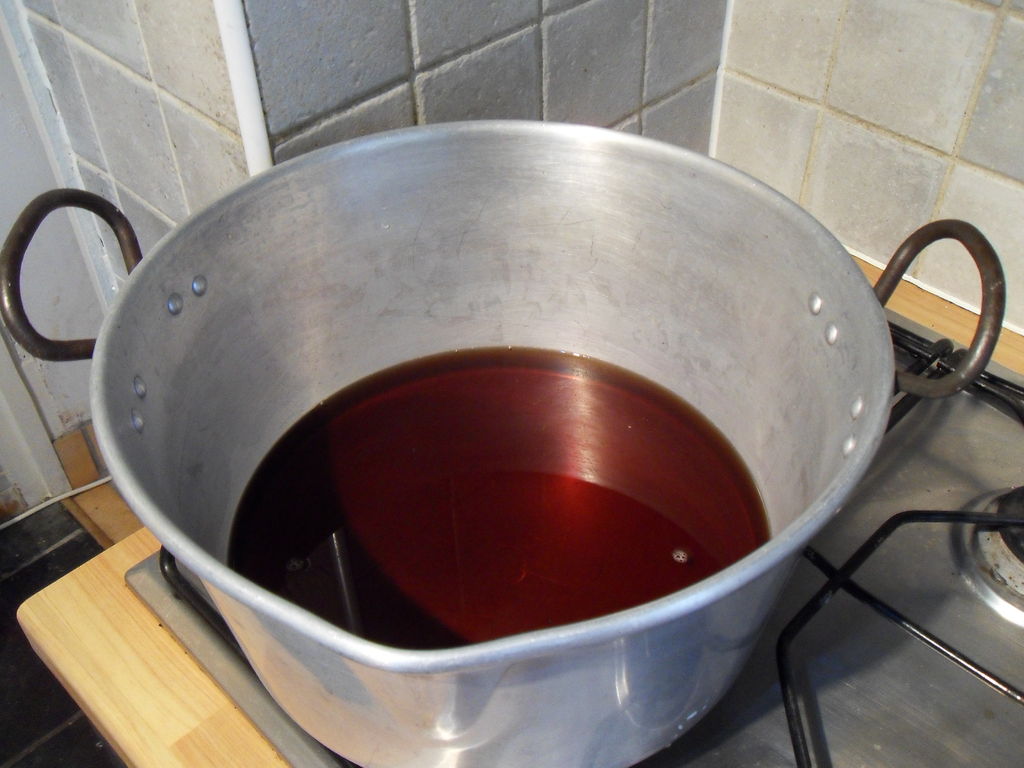}}
     \tcbox[colback=black!85!black, left=0mm,right=0mm,top=0mm,bottom=0mm,boxsep=0mm,boxrule=0.5pt]{\includegraphics[width=0.15\linewidth]{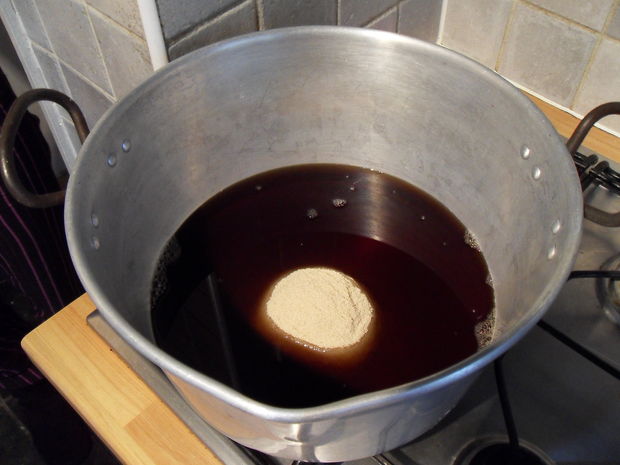}}
     \tcbox[colback=black!85!black, left=0mm,right=0mm,top=0mm,bottom=0mm,boxsep=0mm,boxrule=0.5pt]{\includegraphics[width=0.15\linewidth]{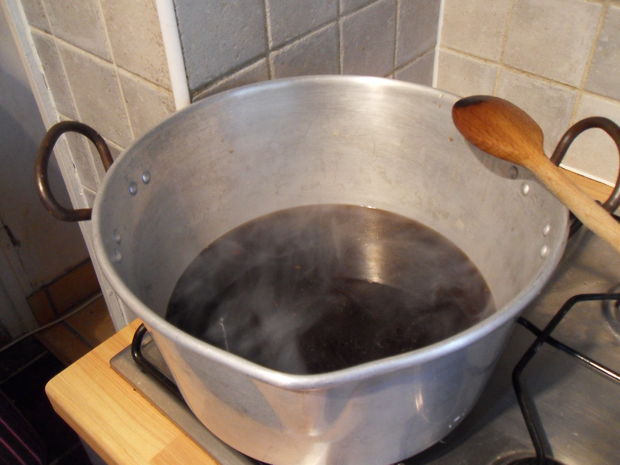}}

     \item[\textbf{Step 4:}] \vspace{-2mm} \dots
     
     %\item[\textbf{Step 5:}] \vspace{-2mm} \dots
     %\item[\vdots] \vspace{-4mm}
     \item[\vdots] \vspace{-4mm}
     
    \item[\textbf{Step 9:}] \vspace{-2mm} While the jars cool, write some labels showing the date, content and maker. Once cool, add the lids and stick on the labels. \dots
    
    \tcbset{nobeforeafter}
     \tcbox[colback=black!85!black, left=0mm,right=0mm,top=0mm,bottom=0mm,boxsep=0mm,boxrule=0.5pt]{\includegraphics[width=0.15\linewidth]{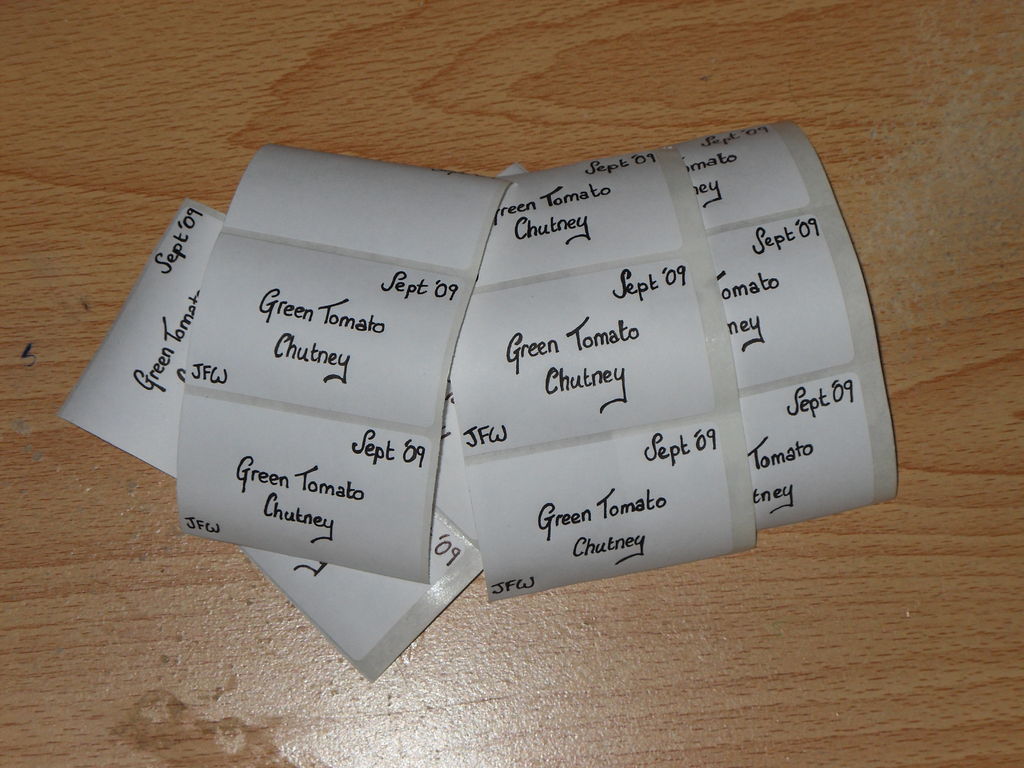}}
     \tcbox[colback=black!85!black, left=0mm,right=0mm,top=0mm,bottom=0mm,boxsep=0mm,boxrule=0.5pt]{\includegraphics[width=0.15\linewidth]{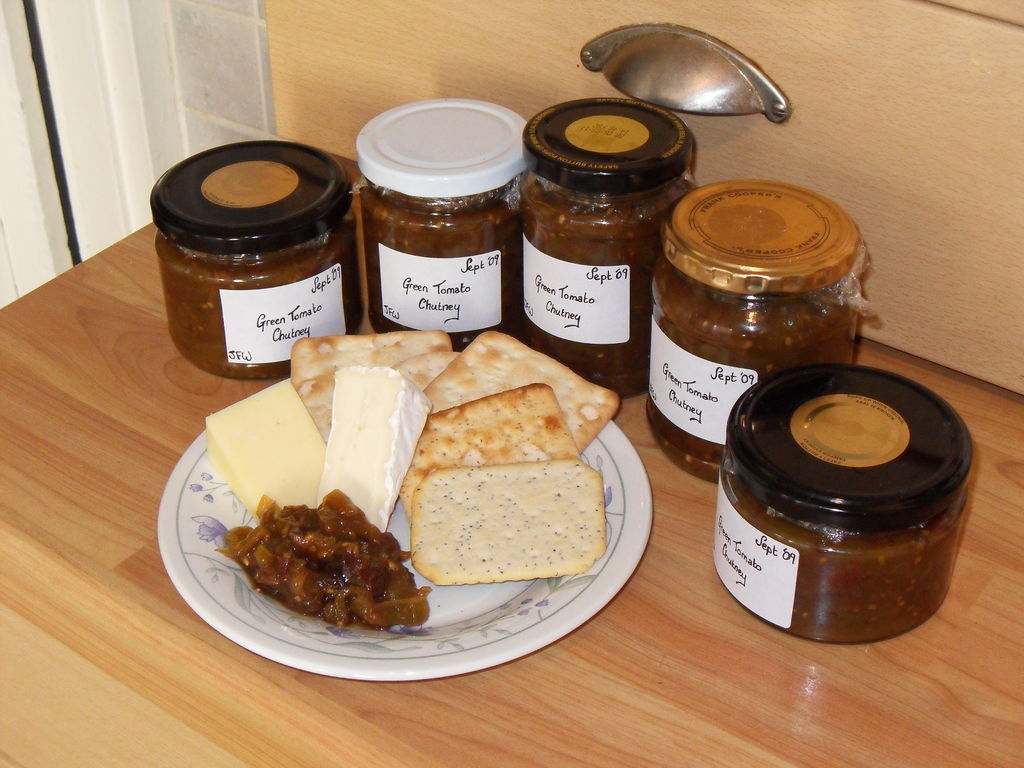}}
    
     \end{enumerate}
    \\
    \midrule
    \multirow{2}{*}{\rotatebox{90}{
        \begin{minipage}{2cm}
        \textbf{Textual Cloze\\Style Question}
        \end{minipage}}}
    & \hspace{0.65cm} \textbf{Question} $\quad$ Choose the best text for the missing blank to correctly complete the recipe \vspace{0.5cm}\\
    & 
    \hspace{2.25cm}
    \tcbset{nobeforeafter}
    %\tcbset{colframe=black!50!black,colback=white,colupper=red!50!black, fonttitle=\bfseries,nobeforeafter,center title}
    Ingredients. \hspace{0.15cm} \underline{\hspace{2cm}}. \hspace{0.15cm} Drain and Add the Tomatoes and Onions. \hspace{0.15cm} Preparing Your Jars.
    % u'Ingredients',
    %           u'@placeholder',
    %           u'Drain and Add the Tomatoes and Onions',
    %           u'Preparing Your Jars'
    
    \\
    & \hspace{0.6cm} \textbf{Answer} $\;\;\;$
    
    \tcbset{nobeforeafter}
    
    \hspace{2.25cm}
    \textcolor{green}{\textbf{A. Sultanas}} \hspace{0.15cm} B. Spicy Tomato Chutney. \hspace{0.15cm} C. Cover and Slice. \hspace{0.15cm} D. Enjoy.

    \hspace{0.1cm}
    \vspace{0.25cm}\\
    & \hspace{0.4cm} 
    \begin{tabular}{ll}
         {Hasty Student:} & \textcolor{red}{Cover and Slice} \\
        {Neural Baseline (Text only):} & \textcolor{green}{Sultanas} \\
         {Neural Baseline (Multimodal):} & \textcolor{green}{Sultanas}
    \end{tabular} 
    \vspace{0.25cm}\\
   
    \bottomrule
  \end{tabularx}
  \caption{\small{Sample groundtruth and model prediction results for a textual cloze style question (context, question and answer triplet) taken from the RecipeQA test set (Question Id: 1000-12665-0-3-4-6). Here, the context is comprised of step descriptions and images where the questions are generated using the step titles in the recipe. The correct answer is in green. The answers selected by the neural models are correct, marked as green whereas Hasty Student's prediction is wrong and marked as red.}}
  \label{fig:textual_cloze_results_supp}
\end{figure*}

%%%%%%%%%%%%%%%%%%%%%%%%%%%
\begin{figure*}[t!]
  \footnotesize
  \begin{tabularx}{\linewidth}{@{}l@{$\;$}Y@{}}
    \toprule
    & \hspace{7.5cm} \textbf{Context Modalities: Titles and Descriptions of Steps} \\
    \midrule
    \multicolumn{2}{@{}l}{\textbf{Recipe: Peppermint-Patty-Pudding-Shot}} \\
    &
    \vspace{-2mm} 
    \begin{enumerate}
     \item[\textbf{Step 1:}] \vspace{-2mm} \textbf{Gather Ingredients}
     To make peppermint patty pudding shots you will need: 1 small box of chocolate pudding3/4 cup of milk3/4 cup of peppermint schnapps1 tub of cool whipCrushed peppermint candy \dots
     \item[\textbf{Step 2:}] \vspace{-2mm} \textbf{Mixing of Ingredients} First wisk together milk and pudding. Once that is combined add in the peppermint schnapps. Then fold in the cool whip.\dots
     \item[\textbf{Step 3:}] \vspace{-2mm} \textbf{Prep for Serving} I then scoop the pudding into small plastic cups with lids. I buy them from a local Chinese restaurant, they are the perfect size. Throw these in the freezer until you are ready to serve. \dots
     \item[\textbf{Step 4:}] \vspace{-2mm} \textbf{Serve} Pull them out of the freezer and sprinkle with the crushed peppermint. You can either lick them out of the cup or eat with a spoon :) I hope you enjoy them as much as we did at Christmas! \dots
     
     \end{enumerate}
    \\
    %------------------------------------
    \midrule
    \multirow{2}{*}{\rotatebox{90}{
        \begin{minipage}{2cm}
        \textbf{Visual Cloze\\Style Question}
        \end{minipage}}}
    & \hspace{0.65cm} \textbf{Question} $\quad$ Choose the best image for the missing blank to correctly complete the recipe \vspace{0.1cm}\\
    & 
    \hspace{2.25cm}
    \tcbset{nobeforeafter}
    %\tcbset{colframe=black!50!black,colback=white,colupper=red!50!black, fonttitle=\bfseries,nobeforeafter,center title}
    \tcbox[colback=black!85!black, left=0mm,right=0mm,top=0mm,bottom=0mm,boxsep=0mm,boxrule=0.5pt]{\includegraphics[width=0.15\linewidth]{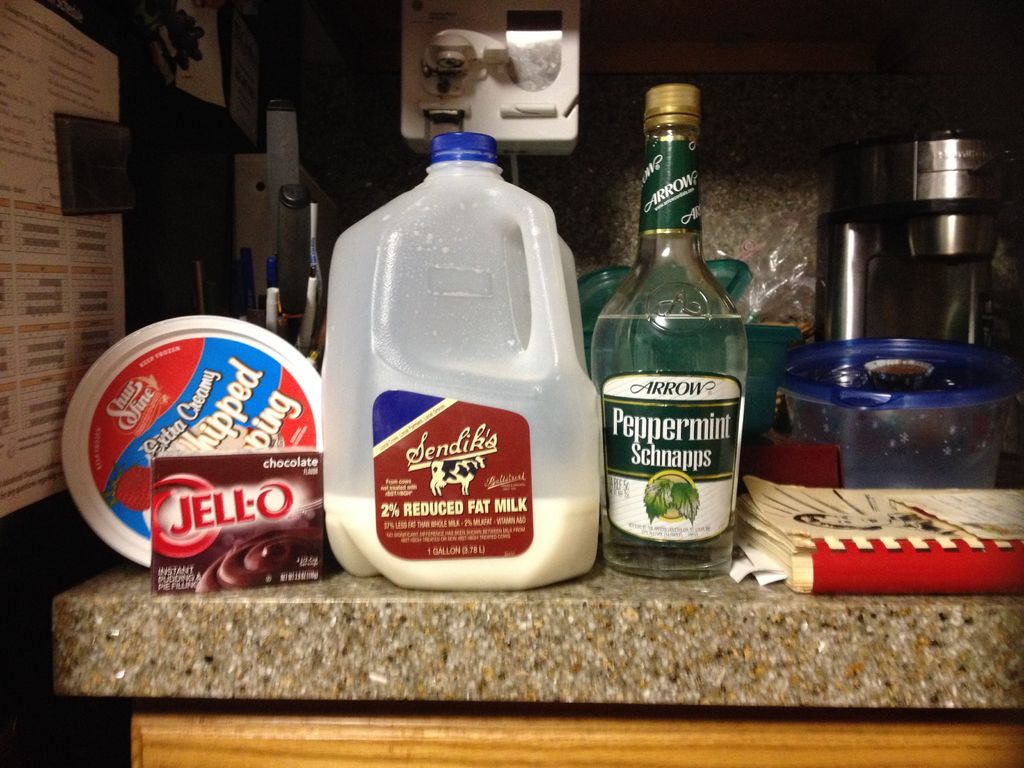}}
    \hspace{0.25cm}
    \tcbox[colback=black!85!black, left=0mm,right=0mm,top=0mm,bottom=0mm,boxsep=0mm,boxrule=0.5pt]{\includegraphics[width=0.15\linewidth]{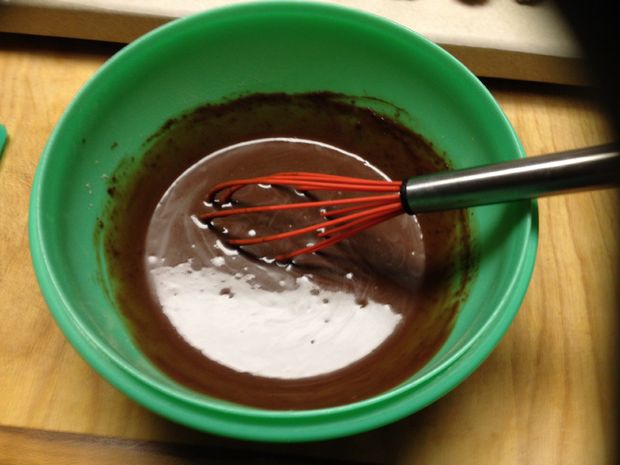}}
    \hspace{0.25cm}
    \tcbox[colback=white!85!white, left=0mm,right=0mm,top=0mm,bottom=0mm,boxsep=0mm,boxrule=0.5pt]{\phantom{\includegraphics[width=0.15\linewidth]{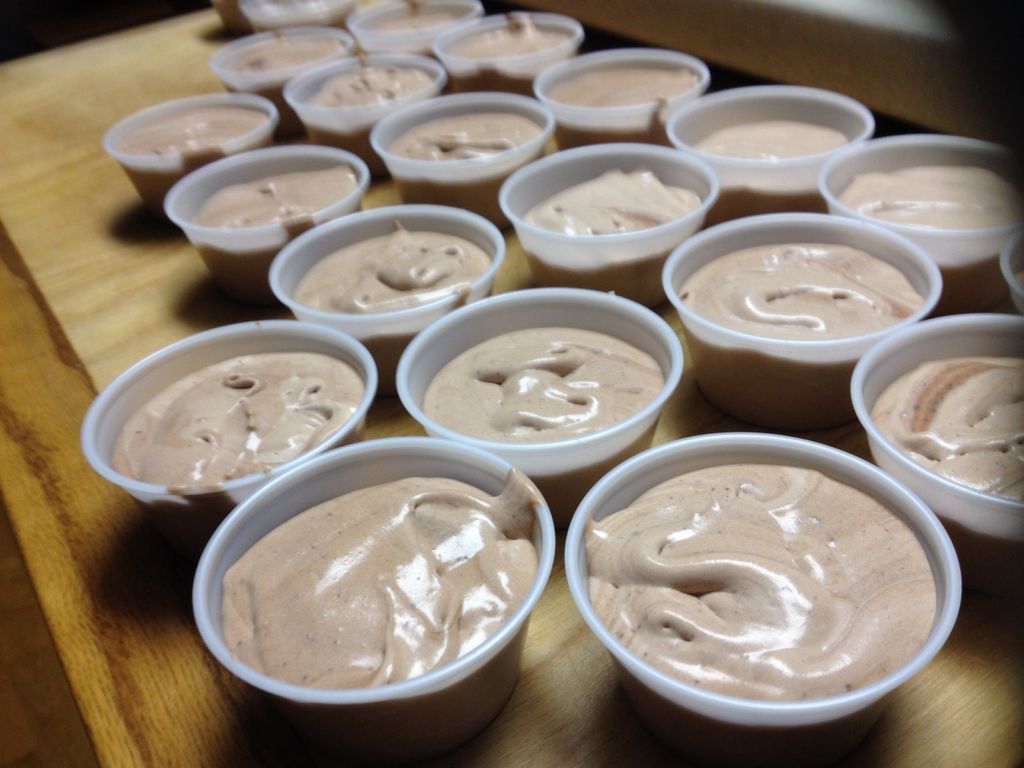}}}
    \hspace{0.25cm}
    \tcbox[colback=black!85!black, left=0mm,right=0mm,top=0mm,bottom=0mm,boxsep=0mm,boxrule=0.5pt]{\includegraphics[width=0.15\linewidth]{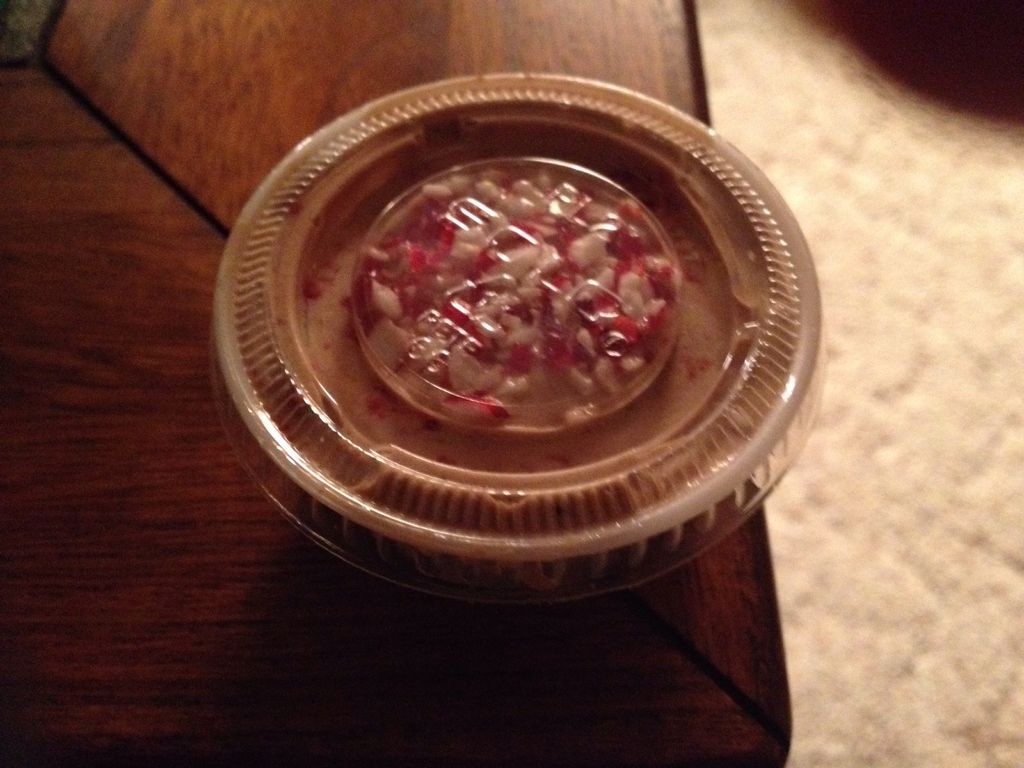}}\vspace{0.25cm}
    \\
    & \hspace{0.8cm} \textbf{Answer} $\;\;\;$
    \tcbset{nobeforeafter}
    %\tcbset{colframe=black!50!black,colback=white,colupper=red!50!black, fonttitle=\bfseries,nobeforeafter,center title}
    \tcbox[colback=black!85!black, left=0mm,right=0mm,top=0mm,bottom=0mm,boxsep=0mm,boxrule=0.5pt]{\includegraphics[width=0.15\linewidth]{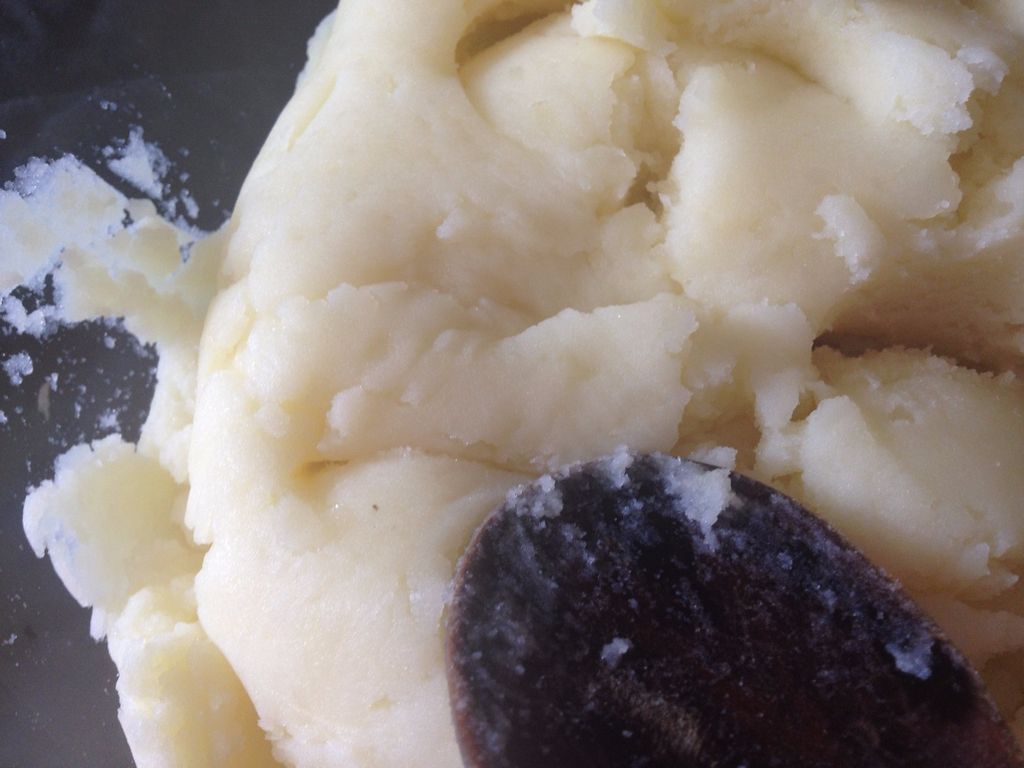}}
    \hspace{0.2cm}
    \tcbox[colback=black!85!black, left=0mm,right=0mm,top=0mm,bottom=0mm,boxsep=0mm,boxrule=0.5pt]{\includegraphics[width=0.15\linewidth]{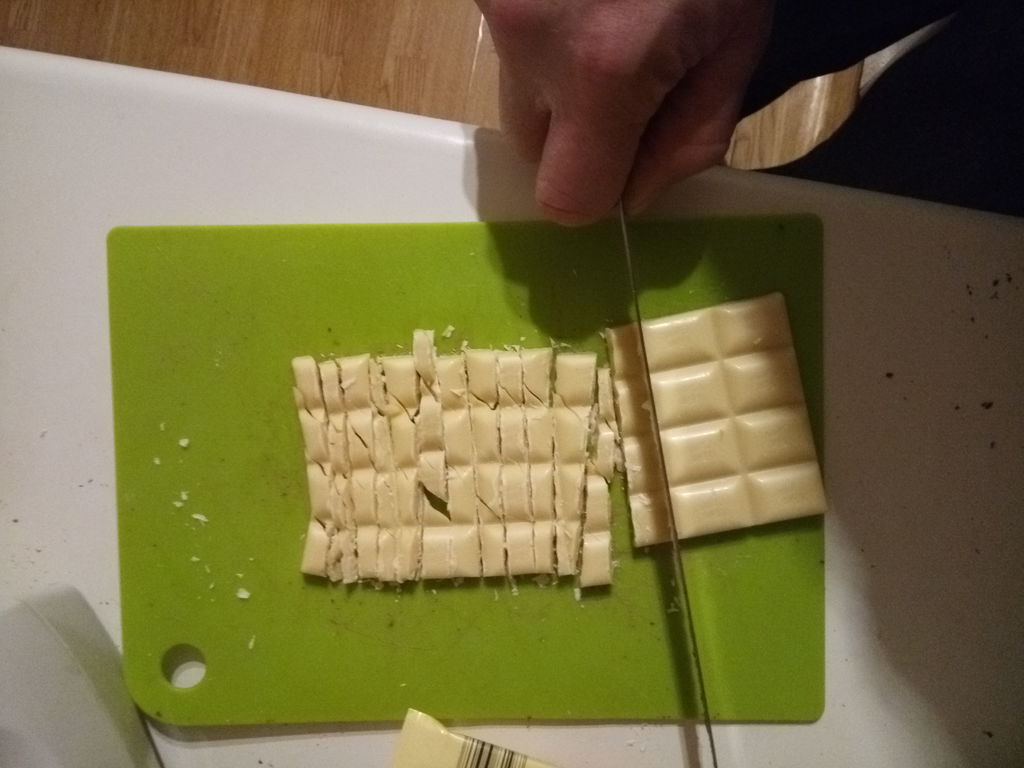}}   
    \hspace{0.25cm}
    \tcbox[colback=black!85!black, left=0mm,right=0mm,top=0mm,bottom=0mm,boxsep=0mm,boxrule=0.5pt]{\includegraphics[width=0.15\linewidth]{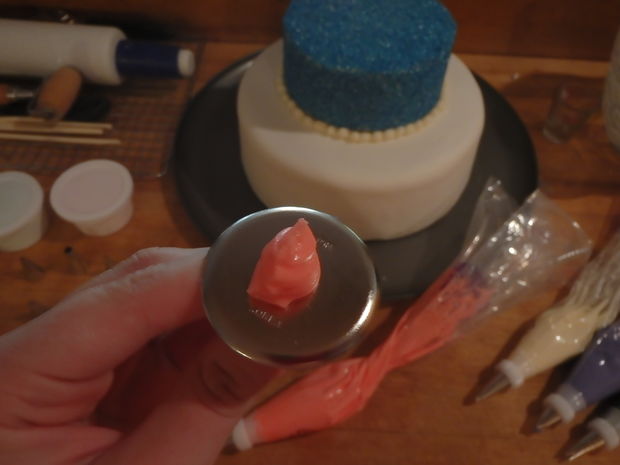}} 
     \hspace{0.2cm}
    \tcbox[colframe=green!30!black,      colback=green!30,left=1mm,right=1mm,top=1mm,bottom=1mm,boxsep=0mm,boxrule=0.5pt]{\includegraphics[width=0.15\linewidth]{img/peppermint-patty-pudding-shot_3_0.jpg}}
    \hspace{0.1cm}\\
     & \hspace{3.4cm} A. \hspace{2.2cm} B. \hspace{2.25cm} C. \hspace{2.4cm} \textcolor{green}{D.} \vspace{0.15cm} \\
    & \hspace{0.4cm} 
    \begin{tabular}{ll}
         {Hasty Student:} & \textcolor{red}{C} \\
         {Neural Baseline (Multimodal):} & \textcolor{red}{C}
    \end{tabular} 
    \vspace{0.25cm}\\
    %------------------------------------
    \midrule
    \multirow{2}{*}{\rotatebox{90}{
        \begin{minipage}{2.4cm}
        \textbf{Visual Coherence\\Style Question}
        \end{minipage}}}
    & \hspace{0.65cm} \textbf{Question} $\quad$ Select the incoherent image in the following sequence of images \vspace{0.1cm}\\
    & \hspace{0.65cm} \textbf{Answer} $\quad\;$
    \tcbset{nobeforeafter}
    %\tcbset{colframe=black!50!black,colback=white,colupper=red!50!black, fonttitle=\bfseries,nobeforeafter,center title}
    \tcbox[colback=white!85!white, left=0mm,right=0mm,top=0mm,bottom=0mm,boxsep=0mm,boxrule=0.5pt]{\includegraphics[width=0.15\linewidth]{img/peppermint-patty-pudding-shot_1_0.jpg}}
    \hspace{0.15cm}
    \tcbox[colback=black!85!black, left=0mm,right=0mm,top=0mm,bottom=0mm,boxsep=0mm,boxrule=0.5pt]{\includegraphics[width=0.15\linewidth,height=0.1125\linewidth]{img/peppermint-patty-pudding-shot_2_2.jpg}}
    \hspace{0.15cm}
    \tcbox[colframe=green!30!black,      colback=green!30,left=1mm,right=1mm,top=1mm,bottom=1mm,boxsep=0mm,boxrule=0.5pt]{\includegraphics[width=0.15\linewidth]{img/potato-cakes_2_3.jpg}}   
    \hspace{0.2cm}
    \tcbox[colback=black!85!black, left=0mm,right=0mm,top=0mm,bottom=0mm,boxsep=0mm,boxrule=0.5pt]{\includegraphics[width=0.15\linewidth]{img/peppermint-patty-pudding-shot_4_0.jpg}}\\
     & \hspace{3.4cm} A. \hspace{2.2cm} B. \hspace{2.25cm} \textcolor{green}{C.} \hspace{2.4cm} D. \vspace{0.25cm} \\
    & \hspace{0.45cm} 
    \begin{tabular}{ll}
         {Hasty Student:} & \textcolor{green}{C} \\
         {Neural Baseline (Multimodal):} & \textcolor{red}{B}
    \end{tabular} 
    \vspace{0.25cm}\\
    \midrule
    \multirow{2}{*}{\rotatebox{90}{
        \begin{minipage}{2.225cm}
        \textbf{Visual Ordering\\Style Question}
        \end{minipage}}}
    & \hspace{0.65cm} \textbf{Question} $\quad$ Choose the correct order of the images to make a complete recipe \vspace{0.1cm}\\
    & 
    \hspace{1.9cm} $\;\;\;$
    \tcbset{nobeforeafter}
    %\tcbset{colframe=black!50!black,colback=white,colupper=red!50!black, fonttitle=\bfseries,nobeforeafter,center title}
    \tcbox[colback=white!85!white, left=0mm,right=0mm,top=0mm,bottom=0mm,boxsep=0mm,boxrule=0.5pt]{\includegraphics[width=0.15\linewidth]{img/peppermint-patty-pudding-shot_4_0.jpg}}
    \hspace{0.25cm}
    \tcbox[colback=black!85!black, left=0mm,right=0mm,top=0mm,bottom=0mm,boxsep=0mm,boxrule=0.5pt]{\includegraphics[width=0.15\linewidth]{img/peppermint-patty-pudding-shot_1_0.jpg}}
    \hspace{0.25cm}
    \tcbox[colback=black!85!black, left=0mm,right=0mm,top=0mm,bottom=0mm,boxsep=0mm,boxrule=0.5pt]{\includegraphics[width=0.15\linewidth]{img/peppermint-patty-pudding-shot_3_0.jpg}}   
    \hspace{0.25cm}
    \tcbox[colback=black!85!black, left=0mm,right=0mm,top=0mm,bottom=0mm,boxsep=0mm,boxrule=0.5pt]{\includegraphics[width=0.15\linewidth]{img/peppermint-patty-pudding-shot_2_2.jpg}}
    \\
    & \hspace{3.45cm} (i) \hspace{2.15cm} (ii) \hspace{2.1cm} (iii) \hspace{2.05cm} (iv) \vspace{0.15cm} \\ 
    & \hspace{0.65cm} \textbf{Answer} $\quad$ A. (i)-(ii)-(iii)-(iv) \hspace{0.25cm} B. (iii)-(i)-(iv)-(ii) \hspace{0.25cm} \textcolor{green}{\textbf{C. (ii)-(iv)-(iii)-(i)}} \hspace{0.25cm} D. (i)-(iii)-(iv)-(ii) \vspace{0.25cm}\\
    &\hspace{0.45cm} 
    \begin{tabular}{ll}
         {Hasty Student:} & \textcolor{red}{B} \\
         {Neural Baseline (Multimodal):} & \textcolor{red}{A}
    \end{tabular} 
    \vspace{0.25cm}\\
    
    \bottomrule
  \end{tabularx}
  \caption{\small{Sample visual cloze, visual coherence and visual ordering style question (context, question and answer triplet) taken from the RecipeQA test set (Question Ids: 2000-13317-0-1-2-3, 3000-13317-0-1-2-3, 4000-13317-0-1-2-3). Here, the context is comprised of step titles and descriptions where the questions are generated using the images in the recipe. The correct answers are shown with green frames or in green. Wrong answers are marked as red.}}
  \label{fig:visualtasks_supp}
\end{figure*}